%% file: main.tex
\documentclass[runningheads]{llncs}

\usepackage{eccv}

\usepackage{eccvabbrv}

\usepackage{graphicx}
\usepackage{booktabs}
\usepackage{amsmath}
\usepackage{amssymb}
\usepackage{multirow}
\usepackage{rotating}
\usepackage{tabularx}
\usepackage{multicol}
\usepackage{makecell}
\usepackage{algorithm}
\usepackage{listings}

\usepackage[accsupp]{axessibility}  

\usepackage{hyperref}

\usepackage{orcidlink}

\begin{document}

\title{3D Single-object Tracking in Point Clouds with High Temporal Variation} 

\titlerunning{3D SOT in Point Clouds with High Temporal Variation}

\author{Qiao Wu\inst{1}\and
Kun Sun\inst{2}\and
Pei An\inst{3}\and
Mathieu Salzmann\inst{4}\and
Yanning Zhang\inst{1}\and
Jiaqi Yang\inst{1}\thanks{Corresponding author.}}

\authorrunning{Q. Wu et al.}


\institute{Northwestern Polytechnical University \and
China University of Geosciences, Wuhan \and
HuaZhong University of Science and Technology \and
École Polytechnique Fédérale de Lausanne \\
\email{qiaowu@mail.nwpu.edu.cn, jqyang@nwpu.edu.cn}}

\maketitle

\begin{abstract}
The high temporal variation of the point clouds is the key challenge of 3D single-object tracking (3D SOT). Existing approaches rely on the assumption that the shape variation of the point clouds and the motion of the objects across neighboring frames are smooth, failing to cope with high temporal variation data. In this paper, we present a novel framework for 3D SOT in point clouds with high temporal variation, called HVTrack. HVTrack proposes three novel components to tackle the challenges in the high temporal variation scenario: 1) A Relative-Pose-Aware Memory module to handle temporal point cloud shape variations; 2) a Base-Expansion Feature Cross-Attention module to deal with similar object distractions in expanded search areas; 3) a Contextual Point Guided Self-Attention module for suppressing heavy background noise. We construct a dataset with high temporal variation (KITTI-HV) by setting different frame intervals for sampling in the KITTI dataset. On the KITTI-HV with 5 frame intervals, our HVTrack surpasses the state-of-the-art tracker CXTracker by \textbf{11.3\%/15.7\%} in Success/Precision.

\end{abstract}

\section{Introduction}
\label{sec:intro}

\begin{figure*}[tp]
  \centering
  \includegraphics[width=1.0\linewidth]{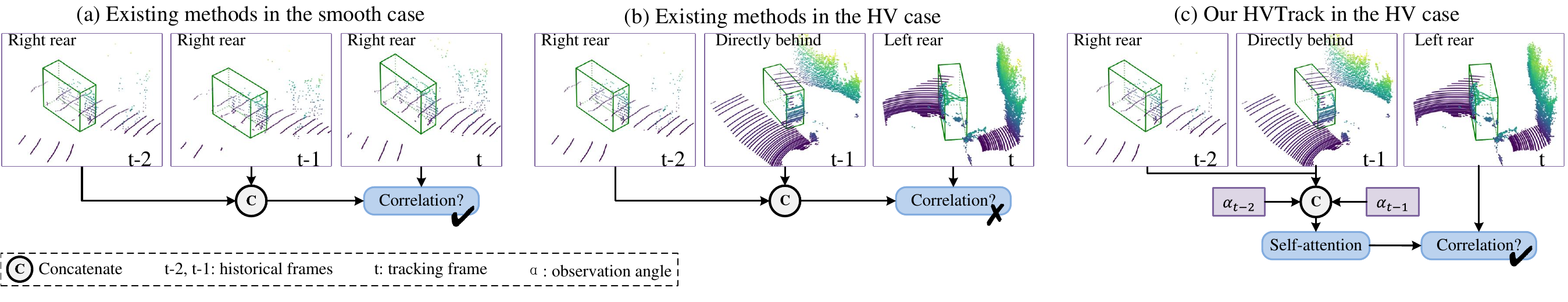}
    \caption{\textbf{Feature correlation in 3D SOT.} \textbf{(a)} Feature correlation in the smooth case (1 frame interval). Correlating the features is relatively trivial as the target undergoes only small shape variations, and the observation angles are consistent in the three frames. \textbf{(b-c)} Feature correlation in high temporal variation cases (10 frames interval). The pose relative to the camera changes rapidly. Correlating the features using historical information is highly challenging (b). We encode the historical observation angles $\alpha $ into the features to guide the variation of relative pose to the camera (c).}
   
  \label{fig:alhpa_motivation}
\end{figure*}

3D single-object tracking (3D SOT) is pivotal for autonomous driving~\cite{yin2021center, cheng2023topology} and robotics~\cite{machida2012human, kart2019object, Ren_2023_HPM, zhang2023mac}. Given the target point cloud and 3D bounding box as template, the goal of 3D SOT is to regress the target 3D poses in the tracking point cloud sequence. Existing approaches~\cite{giancola_leveraging_2019,qi_p2b_2020,fang_3d-siamrpn_2020,hui_3d_2021,hui_3d_2022,wang_mlvsnet_2021,zheng_box-aware_2021,shan_ptt_2021,zheng_beyond_2022,zhou2022pttr,xu2023cxtrack,guo2022cmt,cui20213d} rely on the assumption that the point cloud variations and motion of the object across neighboring frames are relatively smooth. They crop out a small search area around the last proposal for tracking, thus dramatically reducing the complexity of the problem. The template and search area features are then typically correlated as shown in \cref{fig:alhpa_motivation}a, and used to regress the 3D bounding box.

In practice, these approaches are challenged by the presence of large point cloud variations due to the limited sensor temporal resolution and the moving speed of objects as shown in \cref{fig:alhpa_motivation}b. We refer to this significant variation in point cloud and object position between two frames as the high temporal variation (HV).
The high temporal variation challenge is non-negligible in existing benchmarks, and exists in other scenarios not yet covered by them, such as:
\begin{itemize}
\item Skipped-tracking, which can greatly reduce computational consumption in tracking and serve a wide range of other tasks such as detection~\cite{nishimura2022sdof} and segmentation~\cite{yoo2023video}. 
\item Tracking in edge devices, which is essential for deploying trackers on common devices with limited frame rate, resolution, computation, and power \etc.
\item Tracking in highly dynamic scenarios~\cite{kapania2020multi}, which is common in life. For example, tracking in sports events, highway, and UAV scenarios.

\end{itemize}

\begin{figure*}[t]
  \centering
   \includegraphics[width=0.6\linewidth]{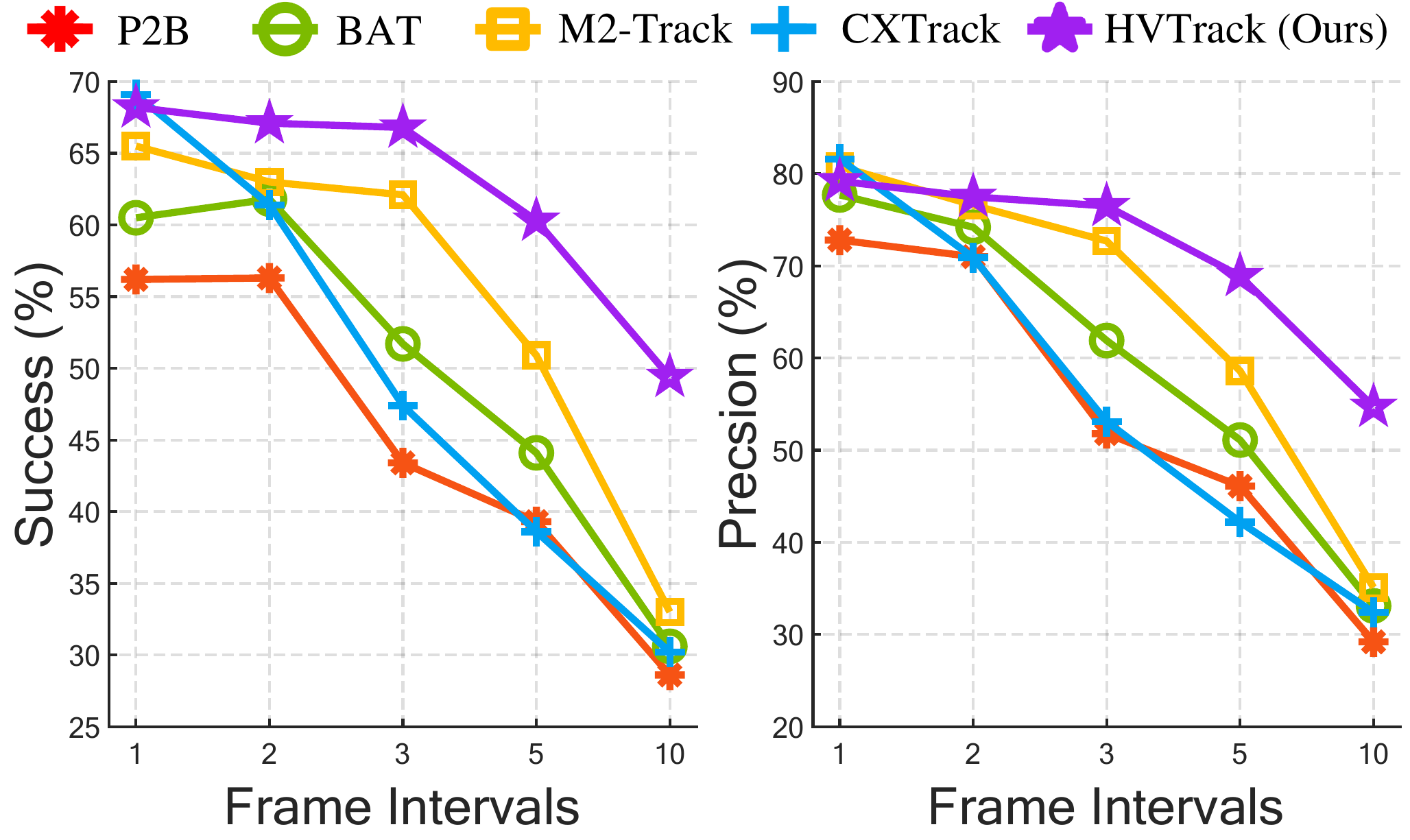}

   \caption{Comparison of HVTrack with the SOTAs~\cite{qi_p2b_2020,zheng_box-aware_2021,zheng_beyond_2022,xu2023cxtrack} on `Car' from KITTI-HV (KITTI~\cite{geiger_are_2012} with different frame intervals, see \cref{sec:experiments}).}
   \label{fig:kitti_hard_fig}
\end{figure*}

 There are three challenges for 3D SOT in HV point clouds, and existing approaches are not sufficient to address these challenges. 1) \emph{Strong shape variations of the point clouds}: Point cloud shape variations are usually caused by the occlusion and relative pose transformation between the object and the sensor. As illustrated in~\cref{fig:alhpa_motivation}b, feature correlation in existing approaches fails because of the dramatic change in the density and distribution of points. 2) \emph{Distractions due to similar objects}: When objects suffer from a significant motion, the search area needs to be enlarged to incorporate the target, thus introducing more distractions from similar objects. Most of the existing trackers focus on local scale features, which discards environmental spatial contextual information to handle distractions. 3) \emph{Heavy background noise}: The expansion of the search area further reduces the proportion of target information in the scene. While aiming to find the high template-response features in the feature correlation stage, existing methods then neglect to suppress the noise interference and reduce the impact of noise features. We evaluate state-of-the-art (SOTA) trackers~\cite{qi_p2b_2020,zheng_box-aware_2021,zheng_beyond_2022,xu2023cxtrack} in the high temporal variation scenario as shown in~\cref{fig:kitti_hard_fig}.
Their performance drops dramatically as the temporal variation of scene point clouds enlarges.

To address the above challenges, we propose a novel framework for 3D SOT in point clouds with \textbf{H}igh temporal \textbf{V}ariation, which we call HVTrack. Specifically, we propose three novel modules to address each of the three above-mentioned challenges. 1) A Relative-Pose-Aware Memory (RPM) module to handle the strong shape variations of the point clouds. Different from~\cite{lan2022temporal}, we integrate the foreground masks and observation angles into the memory bank. Therefore, the model can implicitly learn the distribution variation of point clouds from the relative pose in time. The information arising from observation angles has been overlooked by all existing trackers. 2) A Base-Expansion Feature Cross-Attention (BEA) module to deal with the problem of similar object distractions occurring in large scenes. We synchronize the correlation of the hybrid scales features (base and expansion scales, \cref{sec:BEA}) in the cross-attention, and efficiently utilize spatial contextual information.
3) A Contextual Point Guided Self-Attention (CPA) module to suppress the background noise introduced by the expanded search area. It aggregates the features of points into contextual points according to their importance. 
Less important points share fewer contextual points and vice versa, thus suppressing most of the background noise. BEA and CPA are inspired by the SGFormer~\cite{ren2023sg}, which utilizes hybrid scale significance maps to assign more tokens to salient regions of 2D images. Our experiments clearly demonstrate the remarkable performance of HVTrack in high temporal variation scenarios, as illustrated in~\cref{fig:kitti_hard_fig}. Our contributions can be summarized as follows:
\begin{itemize}

\item For the first time, to the best of our knowledge, we explore the new 3D SOT task for high temporal variation scenarios, and propose a novel framework called HVTrack for the task. 
\item We propose three novel modules, RPM, BEA, and CPA, to address three challenges for 3D SOT in HV point clouds: strong point cloud variations, similar object distractions, and heavy background noise.
\item HVTrack yields state-of-the-art results on KITTI-HV and Waymo, and ranks second on KITTI. Our experimental results demonstrate the robustness of HVTrack in both smooth and high temporal variation cases.

\end{itemize}

\section{Related Work}

\subsection{3D Single-object Tracking} 
Most of the 3D SOT approaches are based on a Siamese framework, because the appearance variations of the target between neighboring frames are not significant. The work of Giancola \etal~\cite{giancola_leveraging_2019} constitutes the pioneering method in 3D SOT. However, it only solved the discriminative feature learning problem, and used a time-consuming and inaccurate heuristic matching to locate the target. Zarzar \etal~\cite{zarzar_efficient_2019} utilized a 2D RPN in bird's eyes view to build an end-to-end tracker. The P2B network~\cite{qi_p2b_2020} employs VoteNet~\cite{qi_deep_2019} as RPN and constructs the first point-based tracker. The following works~\cite{fang_3d-siamrpn_2020,hui_3d_2021,hui_3d_2022,wang_mlvsnet_2021,zheng_box-aware_2021,shan_ptt_2021} develop different architectures of trackers based on P2B~\cite{qi_p2b_2020}. V2B~\cite{hui_3d_2021} leverages the target completion model to generate the dense and complete targets and proposes a simple yet effective voxel-to-BEV target localization network. BAT~\cite{zheng_box-aware_2021} utilizes the relationship between points and the bounding box, integrating the box information into the point clouds. With the development of transformer networks, a number of works~\cite{zhou2022pttr,shan_ptt_2021,hui_3d_2022,guo2022cmt,cui20213d,xu2023cxtrack} have proposed to exploit various attention mechanisms. STNet~\cite{hui_3d_2022} forms an iterative coarse-to-fine cross-and self-attention to correlate the target and search area. CXTrack~\cite{xu2023cxtrack} employs a target-centric transformer to integrate targetness information and contextual information. TAT~\cite{lan2022temporal} leverages the temporal information to integrate target cues by applying an RNN-based~\cite{chung2014empirical} correlation module. Zheng \etal~\cite{zheng_beyond_2022} presented a motion-centric method M2-Track, which is appearance matching-free and has made great progress in dealing with the sparse point cloud tracking problem. Wu \etal~\cite{Wu_2023_MixCycle} proposed the first semi-supervised framework in 3D SOT.

While effective in their context, the above methods are designed based on the assumption that the point cloud variation and motion of the objects across neighboring frames are not significant. In high temporal variation scenarios, this assumption will lead to performance degradation because of the point cloud variations and interference naturally occurring in large scenes. Here, we introduce HVTrack to tackle the challenges of 3D SOT in high temporal variation scenarios. 

\subsection{3D Multi-object Tracking}

3D multi-object tracking (MOT) in point clouds follows two main streams: Tracking-by-detection, and learning-based methods. Tracking-by-detection~\cite{weng20203d,chen2023trajectoryformer, chiu2020probabilistic,wang2021immortal} usually exploits methods such as Kalman filtering to correlate the detection results and track the targets. CenterTrack~\cite{zhou2020tracking}, CenterPoint~\cite{yin2021center}, and SimTrack~\cite{luo2021exploring} replace the filter by leveraging deep networks to predict the velocity and motion of the objects. The learning-based methods~\cite{sadjadpour2023shasta,weng2020gnn3dmot,ding20233dmotformer} typically apply a Graph Neural Network to tackle the association challenge in MOT. GNN3DMOT~\cite{weng2020gnn3dmot} leverages both 2D images and 3D point clouds to obtain a robust association. 3DMOTFormer~\cite{ding20233dmotformer} constructs a graph transformer framework and achieves a good performance using only 3D point clouds.

3D MOT and 3D SOT have different purposes and their own challenges~\cite{jiao2021deep}. 
3D MOT is object-level and focuses on correlating detected objects, whereas 3D SOT is intra-object-level~\cite{jiayao2022real} and aims to track a single object given a template. 3D SOT methods usually come with much lower computational consumption and higher throughput~\cite{zhou2022pttr}. Also, 3D MOT is free from the challenges posed by the dynamic change in the search area size, as MOT is not required to adopt the search area cropping strategy in SOT.

\section{Method}
\subsection{Problem Definition}
Given the template of the target, the goal of 3D SOT is to continually locate the poses of the target in the search area point cloud sequence $\mathbf{P^s} = \{P^{s}_{0},\dots, P^{s}_{t},\dots, P^{s}_{n} | P^{s}_{t} \in \mathbb{R}^{N_{s} \times 3}\}$. Usually, the target point cloud with labels in the first frame is regarded as the template. Former trackers~\cite{giancola_leveraging_2019,qi_p2b_2020,fang_3d-siamrpn_2020,hui_3d_2021,hui_3d_2022,wang_mlvsnet_2021,zheng_box-aware_2021,shan_ptt_2021,zheng_beyond_2022,zhou2022pttr,xu2023cxtrack,guo2022cmt,cui20213d} leverage a 3D bounding box label $B_{0} = (x,y,z,w,l,h,\theta) \in \mathbb{R}^{7}$ to generate the template in the input. Here, $(x,y,z)$, $(w,l,h)$ and $\theta$ are the center location, bounding box size (width, length, and height), and rotation angle of the target, respectively. As objects can be assumed to be rigid, the trackers only need to regress the center and rotation angle of the target.

\subsection{Overview}
\begin{figure*}[t]
  \centering
  \includegraphics[width=0.95\linewidth]{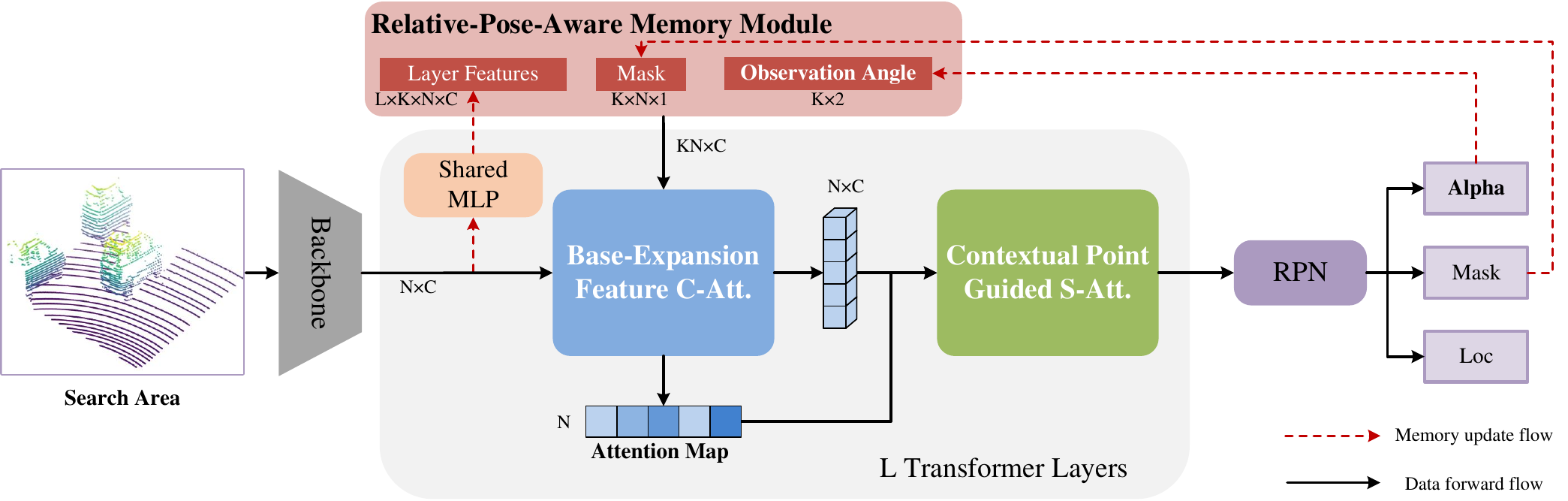}
    \caption{\textbf{HVTrack framework.} We first utilize a backbone to extract the local embedding features of the search area. Then, we construct $L$ transformer layers to fuse spatio-temporal information. For each transformer layer, (i) we apply three memory bank features in the Relative-Pose-Aware Memory module to generate temporal template information; (ii) we employ the Base-Expansion Feature Cross-Attention to correlate the template and search area by leveraging hybrid scale spatial context-aware features; (iii) we introduce a Contextual Point Guided Self-Attention to suppress unimportant noise. After each layer, we update the layer features memory bank using the layer input. Finally, we apply an RPN to regress the 3D bounding box, and update the mask and observation angle memory banks.}
  \label{fig:pipline}
\end{figure*}

We propose HVTrack to exploit both temporal and spatial information and achieve robust tracking in high temporal variation scenarios. As shown in~\cref{fig:pipline}, we take the point cloud $P^{s}_{t}$ at time $t$ as the search area, and leverage memory banks as the template. We first employ a backbone to extract the local spatial features $\mathcal{X}_{0} \in \mathbb{R}^{N \times C}$ of $P^{s}_{t}$, with $N$ and $C$ the point number and feature channel, respectively. Then, $L$ transformer layers are employed to extract spatio-temporal information. For each layer $l$, (i) we capture the template information $Mem_{l} \in \mathbb{R}^{KN \times C}$ from the Relative-Pose-Aware Memory module, with $K$ the memory bank size (\cref{sec:RPM}); (ii) the memory features and search area features $\mathcal{X}_{l-1}$ are correlated in the Base-Expansion Features Cross-Attention (\cref{sec:BEA}); (iii) the Contextual Point Guided Self-Attention (\cref{sec:CPA})  leverages the attention map in the Base-Expansion Features Cross-Attention to suppress the noise features; (iv) we update the Layer Features memory bank using $\mathcal{X}_{l-1}$. After the transformer layers, an RPN is applied to regress the location $(x_{t},y_{t},z_{t}, \theta_{t})$, the mask $\mathcal{M}_{t} \in \mathbb{R}^{N \times 1}$, and the observation angle $\alpha \in \mathbb{R}^{2}$. Finally, the mask and observation angle memory banks are updated using the predicted results.

\subsection{Relative-Pose-Aware Memory Module}
\label{sec:RPM}

As shown in \cref{fig:alhpa_motivation}(b), rapid changes in relative pose lead to large variations in the shape of the object point cloud across the frames. Correlating the object features in $(t-2, t-1, t)$ then becomes difficult, as they have a low overlap with each other. To address this, we introduce the observation angle into the memory bank. The observation angle gives us knowledge of the coarse distribution of an object's point cloud. Thus, the model can learn the variations in point cloud distribution from the historical changes of observation angle.

To exploit the temporal information as the template, we propose a Relative-Pose-Aware Memory (RPM) module. RPM contains 3 memory banks. 1) A layer features memory bank (LM) $\in \mathbb{R}^{L\times K \times N \times C}$: We leverage the historical transformer layer features as the template features to reduce the template inference time in former trackers~\cite{giancola_leveraging_2019,qi_p2b_2020,hui_3d_2021,hui_3d_2022,wang_mlvsnet_2021,zheng_box-aware_2021,shan_ptt_2021,zhou2022pttr,guo2022cmt,cui20213d}. 2) A mask memory bank (MM) $\in \mathbb{R}^{K \times N \times 1}$: Inspired by the mask-based trackers~\cite{zheng_beyond_2022,xu2023cxtrack}, we utilize the mask as the foreground representation. 3) An observation angle memory bank (OM) $\in \mathbb{R}^{K \times 2}$.  For each transformer layer $l$, we process the memory features as
\begin{equation}
    T_{l} = \mathrm{Linear}([\mathrm{LM_{l}},\mathrm{MM}, \mathrm{Repeat}(\mathrm{OM})])\,,
\end{equation}
where $T_{l} \in \mathbb{R}^{KN\times C}$ denotes the template features, $\mathrm{Linear}(\cdot)$ is a linear layer that projects the features from $\mathbb{R}^{KN\times (C+3)}$ to $\mathbb{R}^{KN\times C}$, $[\cdot]$ is the concatenation operation, and $\mathrm{Repeat}(\cdot)$ stacks the OM to $\mathbb{R}^{K\times N\times 2}$. Then, we project $T_{l}$ into Query (Q), Key (K), and Value (V) using the learnable parameter matrices as
\begin{equation}
\begin{aligned}
    Q_{l}^{T} &= \mathrm{LN}(\mathrm{LN}(T_{l})W_{l}^{TQ} + \mathrm{PE}^{T}), \\
    K_{l}^{T} &= \mathrm{LN}(T_{l})W_{l}^{TK},\\
    V_{l}^{T} &= \mathrm{LN}(T_{l})W_{l}^{TV}\,,
\end{aligned}
\end{equation}
where $\mathrm{LN}(\cdot)$ is the layer norm, and $\mathrm{PE}^{T} \in \mathbb{R}^{KN\times C}$ is the positional embedding of the historical point cloud coordinates. We utilize a linear layer to project the point cloud coordinates to their positional embedding. Finally, a self-attention is applied for internal interactions between temporal information as
\begin{equation}
    Mem^{*}_{l} = T_{l} + \mathrm{Dropout}(\mathrm{MHA}(Q_{l}^{T},K_{l}^{T},V_{l}^{T})))\,,
\end{equation}
where MHA is the multi-head attention in \cite{vaswani2017attention}, and $\mathrm{Dropout}$ is the random dropping operation in~\cite{srivastava2014dropout}.
Following CXTrack~\cite{xu2023cxtrack}, we apply dropout and feed-forward network (FFN) after self-attention, i.e.,
\begin{equation}
        Mem_{l} = Mem^{*}_{l} + \mathrm{Dropout}(\mathrm{FFN}(\mathrm{LN}(Mem^{*}_{l})))\,,
\end{equation}
\begin{equation}
    \mathrm{FFN}(x) = \mathrm{max}(0, xW_{1} + b_{1})W_{2} + b_{2}\,.
\end{equation}

\begin{figure}[t]
  \centering
  \begin{subfigure}{0.45\linewidth}
  \centering
  \includegraphics[width=\linewidth]{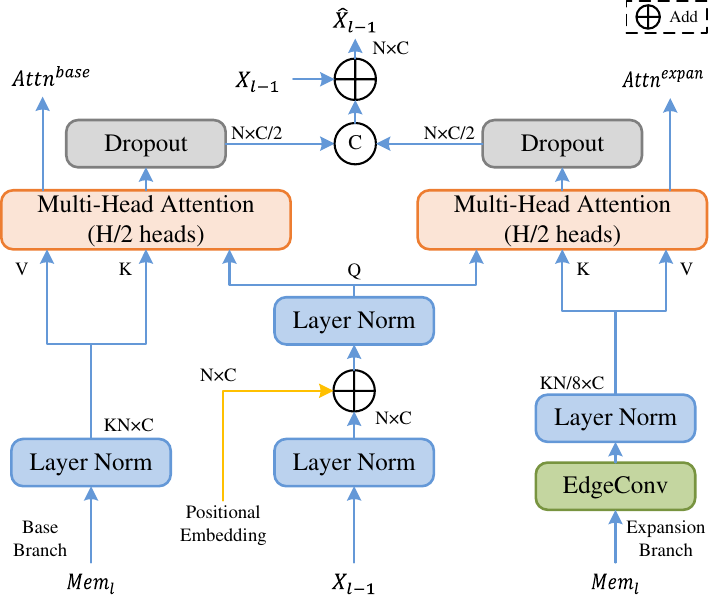}
  \caption{BEA.}
    \label{fig:BEA-a}
  \end{subfigure}
  \hfill
  \begin{subfigure}{0.45\linewidth}
  \centering
  \includegraphics[width=\linewidth]{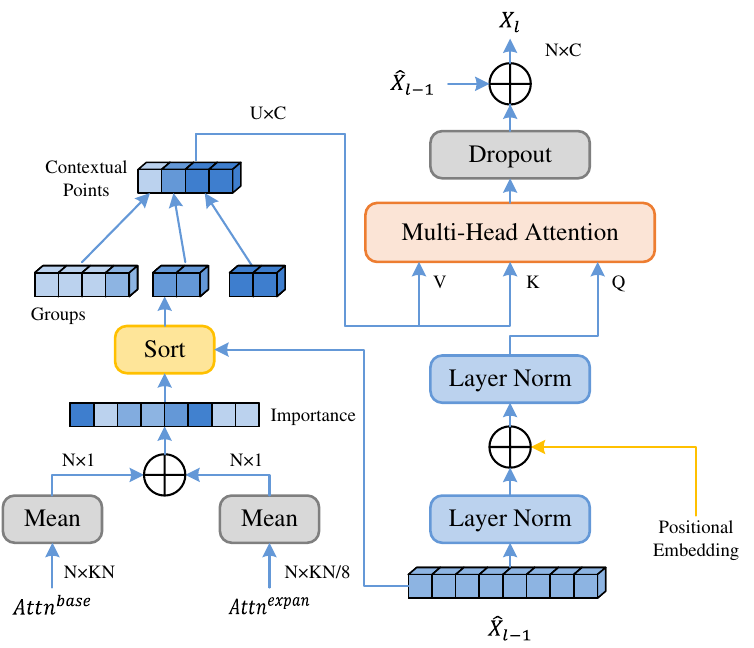}
  \caption{CPA.}
    \label{fig:CPA-a}
  \end{subfigure}
    \caption{\textbf{(a) Base-Expansion Feature Cross-Attention (BEA).} The $H$ heads in the multi-head attention (MHA) are split to process hybrid scale features. For the base scale branch, we directly put the local features into the MHA. For the expansion scale branch, we apply an EdgeConv~\cite{wang2019dynamic} to expand the receptive field of each point and extract more abstract features before MHA. BEA captures the spatial context-aware information with a humble extra computational cost. \textbf{(b) Contextual Point Guided Self-Attention (CPA).} We determine the importance of each point by both base and expansion scale attention maps. Then, we aggregate all the points into $U$ clusters (contextual points) according to their importance and project the clusters to K and V. We assign fewer contextual points for low-importance points, and vice versa. CPA not only suppresses the noise but also reduces the computational cost of the attention.}
  \label{fig:Attns}
\end{figure}

\subsection{Base-Expansion Feature Cross-Attention}
\label{sec:BEA}

Most of the existing trackers~\cite{zheng_box-aware_2021,qi_p2b_2020,zhou2022pttr,shan_ptt_2021,hui_3d_2021,wang_mlvsnet_2021,xu2023cxtrack} employ a point based backbone~\cite{qi2017pointnet++,wang2019dynamic} and focus on local region features, which we call base scale features. Using only base scale features in the whole pipeline is quite efficient and effective in small scenes. However, the base scale features are limited in representing the neighboring environment features around the object in large search areas. To tackle the challenge of similar object distractions, spatial context information across consecutive frames is crucial for effective object tracking~\cite{xu2023cxtrack}. Expanding the receptive field of features can help capture spatial contextual information, and such features are called expansion scale features. Inspired by~\cite{ren2023sg}, we propose Base-Expansion Feature Cross-Attention (BEA) to capture both local and more abstract features, and exploit spatial context-aware information.

As shown in~\cref{fig:BEA-a}, the input features $X_{l-1}$ are projected into Q. Usually, the memory features $Mem_{l}$ would be projected into K and V. Then, multi-head cross-attention adopts $H$ independent heads, and processes them using the same base scale features. By contrast, we split the $H$ heads into $2$ groups. $H/2$ heads exploit local spatial context information. We directly process the base scale features with normal cross-attention, and output base scale features $\hat{X}_{l-1}^{base} \in \mathbb{R}^{N\times C/2}$ and attention map $Attn^{base} \in \mathbb{R}^{N \times KN}$. The other $H/2$ heads capture environment context features. We first apply an EdgeConv~\cite{wang2019dynamic} to extract more abstract features $Mem_{l}^{expan} \in \mathbb{R}^{KN/8 \times C}$, which are expansion scale features, i.e., 
\begin{equation}
    Mem_{l}^{expan} = \mathrm{EdgeConv}(Mem_{l})\,.
\end{equation}
Then, we project the expansion features into K and V, and perform multi-head cross-attention with Q.  Specifically, for the $i$-th head belonging to the expansion scale branch, we generate Q, K, and V as
\begin{equation}
    \begin{aligned}
        Q_{i} &= LN(LN(X_{l-1})W_{i}^{Q} + \mathrm{PE}^{S}_{i}),\\
        K_{i} &= LN(Mem_{l}^{expan})W_{i}^{K},\\
        V_{i} &= LN(Mem_{l}^{expan})W_{i}^{V}\,,
    \end{aligned}
\end{equation}
where  $\mathrm{PE}^{S}_{i}$ is the positional embedding of search area point cloud coordinates. Then, cross-attention is performed as
\begin{equation}
    Attn_{i}^{expan} = \mathrm{Softmax}(\frac{Q_{i}K_{i}}{\sqrt{d_{h}}})\,,
\end{equation}
\begin{equation}
    h_{i}^{expan} = Attn_{i}^{expan}V_{i}\,,
\end{equation}
where $d_{h}$ is the feature dimension of the heads, and $h_{i}^{expan}$ is the output features of the $i$-th head. After that, we concatenate the output features and attention map of each head as
\begin{equation}
    \begin{aligned}
        \hat{X}_{l-1}^{expan} &= [h_{1},\dots, h_{H/2}],\\
        Attn^{expan} &= [Attn_{1}^{expan},\dots,Attn_{H/2}^{expan}]\,,
    \end{aligned}
\end{equation}
where $\hat{X}_{l-1}^{expan} \in \mathbb{R}^{N \times C/2}$, and $Attn^{expan} \in \mathbb{R}^{N \times KN/8}$. Finally, we concatenate the base scale and expansion scale outputs as the resulting correlation feature $\hat{X}_{l-1} \in \mathbb{R}^{N \times C}$. Thus, BEA provides rich hybrid scale spatial contextual information for each point, with a very humble extra computational cost.

\subsection{Contextual Point Guided Self-Attention}
\label{sec:CPA}

Most of the information in the search area will be regarded as noise, because we are only interested in one single object to be tracked. Existing trackers~\cite{zheng_box-aware_2021,qi_p2b_2020,shan_ptt_2021,hui_3d_2021,wang_mlvsnet_2021} aim to find the features with high template-response in the search area, but neglect the suppress to the noise. Zhou \etal~\cite{zhou2022pttr} proposed a Relation-Aware Sampling for preserving more template-relevant points in the search area before inputting it to the backbone. By contrast, we focus on suppressing the noise after feature correlation via a Contextual Point Guided Self-Attention (CPA). 

As shown in~\cref{fig:CPA-a}, we leverage the base and expansion scale attention maps to generate the importance map $I \in \mathbb{R}^{N \times 1}$ as 
\begin{equation}
    I = \mathrm{Mean}(Attn^{base}) + \mathrm{Mean}(Attn^{expan})\,.
\end{equation}
The higher the importance of the point, the more spatial context-aware information related to the target it contains. We sort the points according to the magnitude of their importance values. Then, all the points will be separated into $G$ groups according to their importance. For each group with points $P_{i}^{G} \in \mathbb{R}^{G_{i} \times C}$, we aggregate the points into $U_{i}$ clusters, which we call contextual points. Specifically, we first reshape the points as $P_{i}^{G} \in \mathbb{R}^{U_{i} \times C \times G_{i}/U_{i}}$. Second, a linear layer is employed to project the group to the contextual points $P_{i}^{U} \in \mathbb{R}^{U_{i} \times C}$. We assign fewer contextual points for the groups with lower importance, and suppress the noise feature expression. Finally, all the contextual points are concatenated and projected into Key $K^{U} \in \mathbb{R}^{U\times C}$ and Value $V^{U} \in \mathbb{R}^{U\times C}$. We project $\hat{X}_{l-1}$ to Q and perform a multi-head attention with $K^{U}$ and $V^{U}$,
and an FFN is applied after attention. CPA shrinks the length of K and V, and leads to a computational cost decrease in self-attention.

\subsection{Implementation Details}
\noindent\textbf{Backbone \& Loss Functions.} Following CXTrack~\cite{xu2023cxtrack}, we adopt DGCNN~\cite{wang2019dynamic} as our backbone, and apply X-RPN~\cite{xu2023cxtrack} as the RPN of our framework. We add two Shared MLP layers to X-RPN for predicting the observation angles ($\alpha$) and the masks. Therefore, the overall loss is expressed as
\begin{equation}
    \mathcal{L} = \gamma_{1}\mathcal{L}_{cc} + \gamma_{2}\mathcal{L}_{mask} + \gamma_{3}\mathcal{L}_{alpha} + \gamma_{4}\mathcal{L}_{rm} + \gamma_{5}\mathcal{L}_{box}\,,
    \label{eq:loss_fc}
\end{equation}
where $\mathcal{L}_{cc}$, $\mathcal{L}_{mask}$, $\mathcal{L}_{alpha}$, $\mathcal{L}_{box}$, and $\mathcal{L}_{box}$ are the loss for the coarse center, foreground mask, observation angle, targetness mask, and bounding box, respectively. We apply the $L_{2}$ loss for $\mathcal{L}_{cc}$, the standard cross entropy loss for $\mathcal{L}_{mask}$ and $\mathcal{L}_{rm}$, and the Huber loss for $\mathcal{L}_{alpha}$ and $\mathcal{L}_{box}$. $\gamma_{1}$, $\gamma_{2}$, $\gamma_{3}$, $\gamma_{4}$, and $\gamma_{5}$ are empirically set as $10.0$, $0.2$, $1.0$, $1.0$, and $1.0$. 

\noindent\textbf{Training \& Testing.} We train our model on NVIDIA RTX-3090 GPUs with the Adam optimizer and an initial learning rate of $0.001$. Due to GPU memory limitation, we construct point cloud sequences with $8$ frames for training, and set $K = 2$ in training, and $K = 6$ in testing. Following existing methods~\cite{zheng_beyond_2022,xu2023cxtrack}, we set $N$ and $C$ to 128. We stack $L=2$ transformer layers and apply $H=4$ heads in BEA and CPA. We adopt $G=3$ groups in CPA, and assign $[32,64,32]$ points and $U = [4,32,16]$ contextual points for the groups, respectively.

\section{Experiments}
\label{sec:experiments}
We leverage two famous 3D tracking benchmarks of KITTI~\cite{geiger_are_2012} and Waymo~\cite{sun_scalability_2020} to evaluate the general performance of our approach in regular 3D SOT. In addition, we establish a new KITTI-HV dataset to test our performance in high temporal variation scenarios.

\noindent \textbf{Regular Datasets.} The KITTI tracking dataset comprises $21$ training sequences and $29$ test sequences, encompassing eight object types. Following prior studies~\cite{giancola_leveraging_2019,qi_p2b_2020,zheng_box-aware_2021,wang_mlvsnet_2021,xu2023cxtrack,zheng_beyond_2022}, we use the sequences $0$-$16$ as training data, $17$-$18$ for validation, and $19$-$20$ for testing. The Waymo dataset is large-scale. We adopt the approach outlined in LiDAR-SOT~\cite{pang2021model} to utilize $1121$ tracklets, which are subsequently categorized into easy, medium, and hard subsets based on the number of points in the first frame of each tracklet.

\noindent \textbf{HV Dataset.} We build a dataset with high temporal variation for 3D SOT based on KITTI, called KITTI-HV. Although high temporal variation scenarios are present in the existing benchmarks, there is no exact threshold to determine whether the scenario is a high temporal variation scenario or not. Large point cloud variations and significant object motions are two major challenges in high temporal variation scenarios. Sampling at frame intervals is a good way to simulate these two challenges. Also, the constructed KITTI-HV can provide a preliminary platform for exploring tracking in scenarios such as skipped-tracking, edge devices, and high dynamics. For a fairer comparison with existing methods, we set the frame interval to $2$, $3$, $5$, and $10$. We set up more dense testings at low frame intervals to exploit the performance of the existing methods in point cloud variations close to smooth scenarios. We train and test all methods from scratch individually on each frame interval.

\noindent \textbf{Evaluation Metrics.} We employ One Pass Evaluation~\cite{wu_online_2013} to evaluate the different methods in terms of Success and Precision. Success is determined by measuring the Intersection Over Union between the proposed bounding box and the ground-truth (GT) bounding box. Precision is evaluated by computing the Area Under the Curve of the distance error between the centers of the two bounding boxes, ranging from 0 to 2 meters.

\begin{table*}[ht]
  \centering
  \caption{\textbf{Comparison of HVTrack with the state-of-the-art methods on each category of the KITTI-HV dataset}. We construct the HV dataset KITTI-HV for training and testing by setting different frame intervals for sampling in the KITTI dataset. \textbf{Bold} and \underline{underline} denote the best and second-best performance, respectively. Success/Precision are used for evaluation. Improvement and deterioration are shown in \textcolor[rgb]{ 0,  .69,  .314}{green} and \textcolor[rgb]{ 1,  0,  0}{red}, respectively.}
  \resizebox{\linewidth}{!}{
  \begin{tabular}{c|ccccc|ccccc}
    \toprule[1.5pt]
    Frame Intervals & \multicolumn{5}{c|}{2 Intervals}      & \multicolumn{5}{c}{3 Intervals} \\
    \midrule
    Category & Car   & Pestrian & Van   & Cyclist & Mean  & Car   & Pestrian & Van   & Cyclist & Mean \\
    Frame Number & 6424  & 6088  & 1248  & 308   & 14068 & 6424  & 6088  & 1248  & 308   & 14068 \\
    \midrule
    P2B\cite{qi_p2b_2020}   & 56.3/71.0 & 30.8/53.0 & 33.4/38.4 & 41.8/61.4 & 42.9/60.1 & 43.4/51.8 & 27.9/46.8 & 27.9/31.8 & 44.8/64.4 & 35.4/48.1 \\
    BAT~\cite{zheng_box-aware_2021}   & 61.8/74.2 & 36.5/61.1 & 26.8/30.4 & 54.1/78.7 & 47.6/64.7 & 51.7/61.9 & 31.8/53.5 & 24.0/28.2 & 50.5/72.6 & 40.6/55.5 \\
    M2-Track~\cite{zheng_beyond_2022} & \underline{63.0}/\underline{76.6} & 54.6/81.7 & \underline{52.8}/\underline{66.5} & \underline{68.3}/\underline{89.3} & 58.6/\underline{78.2} & \underline{62.1}/\underline{72.7} & \underline{51.8}/\underline{74.3} & 33.6/41.6 & \underline{64.7}/\underline{82.0} & \underline{55.1}/\underline{70.8} \\
    CXTrack~\cite{xu2023cxtrack} & 61.4/70.9 & \textbf{62.6}/\textbf{86.3} & \textbf{56.0}/\textbf{69.1} & 59.2/76.9 & \underline{61.4}/77.5 & 47.4/53.1 & \textbf{57.9}/\textbf{79.3} & \textbf{48.5}/\textbf{58.8} & 40.7/58.4 & 51.9/65.1 \\
    \midrule
    HVTrack & \textbf{67.1}/\textbf{77.5} & \underline{60.0}/\underline{84.0} & 50.6/61.7 & \textbf{73.9}/\textbf{93.6} & \textbf{62.7}/\textbf{79.3} & \textbf{66.8}/\textbf{76.5} & 51.1/71.9 & \underline{38.7}/\underline{46.9} & \textbf{66.5}/\textbf{89.7} & \textbf{57.5}/\textbf{72.2} \\
    Improvement & \textcolor[rgb]{ 0,  .69,  .314}{4.1↑}/\textcolor[rgb]{ 0,  .69,  .314}{0.9↑} & \textcolor[rgb]{ 1,  0,  0}{2.6↓}/\textcolor[rgb]{ 1,  0,  0}{2.3↓} & \textcolor[rgb]{ 1,  0,  0}{6.0↓/7.4↓} & \textcolor[rgb]{ 0,  .69,  .314}{5.6↑}/\textcolor[rgb]{ 0,  .69,  .314}{4.3↑} & \textcolor[rgb]{ 0,  .69,  .314}{1.3↑}/\textcolor[rgb]{ 0,  .69,  .314}{1.1↑} & \textcolor[rgb]{ 0,  .69,  .314}{4.7↑}/\textcolor[rgb]{ 0,  .69,  .314}{3.8↑} & \textcolor[rgb]{ 1,  0,  0}{6.8↓}/\textcolor[rgb]{ 1,  0,  0}{7.4↓} & \textcolor[rgb]{ 1,  0,  0}{9.8↓}/\textcolor[rgb]{ 1,  0,  0}{11.9↓} & \textcolor[rgb]{ 0,  .69,  .314}{1.8↑}/\textcolor[rgb]{ 0,  .69,  .314}{7.7↑} & \textcolor[rgb]{ 0,  .69,  .314}{2.4↑}/\textcolor[rgb]{ 0,  .69,  .314}{1.4↑} \\
    \midrule
    \midrule
    Frame Intervals & \multicolumn{5}{c|}{5 Intervals}      & \multicolumn{5}{c}{10 Intervals} \\
    \midrule
    Category & Car   & Pestrian & Van   & Cyclist & Mean  & Car   & Pestrian & Van   & Cyclist & Mean \\
    Frame Number & 6424  & 6088  & 1248  & 308   & 14068 & 6424  & 6088  & 1248  & 308   & 14068 \\
    \midrule
    P2B~\cite{qi_p2b_2020}   & 39.3/46.1 & 27.4/43.5 & 27.2/30.4 & 35.0/44.4 & 33.0/43.5 & 28.6/29.2 & \textbf{23.1}/\textbf{31.1} & \textbf{25.9}/\textbf{27.3} & 29.1/28.3 & \underline{26.0}/29.8 \\
    BAT~\cite{zheng_box-aware_2021}   & 44.1/51.1 & 21.1/32.8 & 26.1/29.5 & 35.7/46.3 & 32.4/41.1 & 30.6/33.1 & 21.7/\underline{29.2} & 20.8/20.7 & \underline{29.3}/\underline{29.1} & 25.9/\underline{30.2} \\
    M2-Track~\cite{zheng_beyond_2022} & \underline{50.9}/\underline{58.6} & 31.6/45.4 & \textbf{30.0}/\textbf{36.5} & \underline{47.4}/\underline{61.0} & \underline{40.6}/\underline{51.0} & \underline{33.0}/\underline{35.1} & 17.5/24.1 & 20.7/20.8 & 27.7/26.6 & 25.0/28.9 \\
    CXTrack~\cite{xu2023cxtrack} & 38.6/42.2 & \underline{35.0}/\underline{47.8} & 21.6/24.3 & 25.7/33.3 & 35.3/42.8 & 30.2/32.4 & 18.2/21.4 & 17.5/17.9 & 27.7/26.5 & 23.8/26.2 \\
    \midrule
    HVTrack & \textbf{60.3}/\textbf{68.9} & \textbf{35.1}/\textbf{52.1} & \underline{28.7}/\underline{32.4} & \textbf{58.2}/\textbf{71.7} & \textbf{46.6}/\textbf{58.5} & \textbf{49.4}/\textbf{54.7} & \underline{22.5}/29.1 & \underline{22.2}/\underline{23.4} & \textbf{39.5}/\textbf{45.4} & \textbf{35.1}/\textbf{40.6} \\
    Improvement & \textcolor[rgb]{ 0,  .69,  .314}{9.4↑}/\textcolor[rgb]{ 0,  .69,  .314}{10.3↑} & \textcolor[rgb]{ 0,  .69,  .314}{0.1↑}/\textcolor[rgb]{ 0,  .69,  .314}{4.3↑} & \textcolor[rgb]{ 1,  0,  0}{1.3↓}/\textcolor[rgb]{ 1,  0,  0}{4.1↓} & \textcolor[rgb]{ 0,  .69,  .314}{10.8↑}/\textcolor[rgb]{ 0,  .69,  .314}{10.7↑} & \textcolor[rgb]{ 0,  .69,  .314}{6.0↑}/\textcolor[rgb]{ 0,  .69,  .314}{7.5↑} & \textcolor[rgb]{ 0,  .69,  .314}{16.4↑}/\textcolor[rgb]{ 0,  .69,  .314}{19.6↑} & \textcolor[rgb]{ 1,  0,  0}{0.6↓}/\textcolor[rgb]{ 1,  0,  0}{0.1↓} & \textcolor[rgb]{ 1,  0,  0}{3.7↓}/\textcolor[rgb]{ 1,  0,  0}{3.9↓} & \textcolor[rgb]{ 0,  .69,  .314}{10.2↑}/\textcolor[rgb]{ 0,  .69,  .314}{16.3↑} & \textcolor[rgb]{ 0,  .69,  .314}{9.1↑}/\textcolor[rgb]{ 0,  .69,  .314}{10.4} \\
    \bottomrule[1.5pt]
    \end{tabular}%
    }

  \label{tab:KITTI_Hard}%
\end{table*}%

\begin{table}[tbp]
  \centering
  \caption{Comparison of HVTrack with the SOTA methods on each category of the KITTI dataset.}
  \resizebox{0.55\linewidth}{!}{
    \begin{tabular}{c|c|c|c|c|c}
    \toprule[1.5pt]
    Category & Car   & Pedestrian & Van   & Cyclist & Mean \\
    Frame Number & 6424  & 6088  & 1248  & 308   & 14068 \\
    \midrule
    SC3D~\cite{giancola_leveraging_2019}  & 41.3/57.9 & 18.2/37.8 & 40.4/47.0 & 41.5/70.4 & 31.2/48.5 \\
    P2B~\cite{qi_p2b_2020}   & 56.2/72.8 & 28.7/49.6 & 40.8/48.4 & 32.1/44.7 & 42.4/60.0 \\
    3DSiamRPN~\cite{fang_3d-siamrpn_2020} & 58.2/76.2 & 35.2/56.2 & 45.7/52.9 & 36.2/49.0 & 46.7/64.9 \\
    MLVSNet~\cite{wang_mlvsnet_2021} & 56.0/74.0 & 34.1/61.1 & 52.0/61.4 & 34.3/44.5 & 45.7/66.7 \\
    BAT~\cite{zheng_box-aware_2021}   & 60.5/77.7 & 42.1/70.1 & 52.4/67.0 & 33.7/45.4 & 51.2/72.8 \\
    PTT~\cite{shan_ptt_2021}   & 67.8/81.8 & 44.9/72.0 & 43.6/52.5 & 37.2/47.3 & 55.1/74.2 \\
    V2B~\cite{hui_3d_2021}   & 70.5/81.3 & 48.3/73.5 & 50.1/58.0 & 40.8/49.7 & 58.4/75.2 \\
    PTTR~\cite{zhou2022pttr}  & 65.2/77.4 & 50.9/81.6 & 52.5/61.8 & 65.1/90.5 & 57.9/78.1 \\
    STNet~\cite{hui_3d_2022} & \underline{72.1}/\textbf{84.0} & 49.9/77.2 & 58.0/70.6 & 73.5/93.7 & 61.3/80.1 \\
    TAT~\cite{lan2022temporal}  & \textbf{72.2}/\underline{83.3} & 57.4/84.4 & \underline{58.9}/69.2 & \textbf{74.2}/\underline{93.9} & 64.7/82.8 \\
    M2-Track~\cite{zheng_beyond_2022} & 65.5/80.8 & 61.5/88.2 & 53.8/\underline{70.7} & 73.2/93.5 & 62.9/\underline{83.4} \\
    CXTrack~\cite{xu2023cxtrack} & 69.1/81.6 & \textbf{67.0}/\textbf{91.5} & \textbf{60.0}/\textbf{71.8} & \textbf{74.2}/\textbf{94.3} & \textbf{67.5}/\textbf{85.3} \\
    \midrule
    HVTrack & 68.2/79.2 & \underline{64.6}/\underline{90.6} & 54.8/63.8 & 72.4/93.7 & \underline{65.5}/83.1 \\
    \bottomrule[1.5pt]
    \end{tabular}%
    }
    
  \label{tab:KITTI}%
\end{table}%

\begin{table*}[ht]
  \centering
  \caption{Comparison of HVTrack with the SOTA methods on the Waymo dataset.}
   \resizebox{0.95\linewidth}{!}{
    \begin{tabular}{c|cccc|cccc|c}
    \toprule[1.5pt]
    \multirow{2}[2]{*}{Method} & \multicolumn{4}{c|}{Vehicle (185632)} & \multicolumn{4}{c|}{Pedestrian (241168)} & \multirow{2}[2]{*}{Mean (426800)} \\
          & Easy  & Medium & Hard  & Mean  & Easy  & Medium & Hard  & Mean  &  \\
    \midrule
    P2B~\cite{qi_p2b_2020}   & 57.1/65.4 & 52.0/60.7 & 47.9/58.5  & 52.6/61.7 & 18.1/30.8 & 17.8/30.0 & 17.7/29.3 & 17.9/30.1 & 33.0/43.8 \\
    BAT~\cite{zheng_box-aware_2021}   & 61.0/68.3 & 53.3/60.9  & 48.9/57.8 & 54.7/62.7 & 19.3/32.6 & 17.8/29.8 & 17.2/28.3 & 18.2/30.3 & 34.1/44.4 \\
    V2B~\cite{hui_3d_2021}   & 64.5/71.5 & 55.1/63.2 & 52.0/62.0 & 57.6/65.9 & 27.9/43.9 & 22.5/36.2 & 20.1/33.1 & 23.7/37.9 & 38.4/50.1 \\
    STNet~\cite{hui_3d_2022} & \underline{65.9}/\underline{72.7} & \textbf{57.5}/\textbf{66.0} & \underline{54.6}/\underline{64.7} & \underline{59.7}/\underline{68.0} & 29.2/45.3 & 24.7/38.2 & 22.2/35.8 & 25.5/39.9 & 40.4/52.1 \\
    CXTrack~\cite{xu2023cxtrack} & 63.9/71.1 & 54.2/62.7 & 52.1/63.7 & 57.1/66.1 & \textbf{35.4}/\textbf{55.3} & \textbf{29.7}/\textbf{47.9} & \underline{26.3}/\underline{44.4} & \textbf{30.7}/\textbf{49.4} & \underline{42.2}/\underline{56.7} \\
    \midrule
    HVTrack(Ours) & \textbf{66.2}/\textbf{75.2} & \underline{57.0}/\textbf{66.0} & \textbf{55.3}/\textbf{67.1} & \textbf{59.8}/\textbf{69.7} & \underline{34.2}/\underline{53.5} & \underline{28.7}/\textbf{47.9} & \textbf{26.7}/\textbf{45.2} & \underline{30.0}/\underline{49.1} & \textbf{43.0}/\textbf{58.1} \\
    \bottomrule[1.5pt]
    \end{tabular}%
    }
  \label{tab:Waymo}%
\end{table*}%

\subsection{Comparison with the State of the Art}

\noindent\textbf{Results on HV tracking.}
We evaluate our HVTrack in 4 categories (`Car', `Pedestrian', `Van', and `Cyclist') following existing methods~\cite{qi_p2b_2020,zheng_box-aware_2021,zheng_beyond_2022,xu2023cxtrack} in the KITTI-HV dataset. \emph{The methods we choose to compare with HVTrack are the most representative SOT methods from 2020 to 2023 (Most cited methods published in each year according to Google Scholar).} As illustrated in \cref{tab:KITTI_Hard}, our approach consistently outperforms the state-of-the-art methods~\cite{qi_p2b_2020,zheng_box-aware_2021,zheng_beyond_2022,xu2023cxtrack} across all frame intervals, confirming the effectiveness of the proposed tracking framework for high temporal variation scenarios. Notably, the performance gap between our HVTrack and existing trackers widens as variations are exacerbated. In the particularly challenging scenario of 10 frame intervals, we achieve a substantial \textbf{9.1\%↑} improvement in success and a remarkable \textbf{10.4\%↑} enhancement in precision. This showcases the robustness of our method in accommodating various levels of point cloud variation. Our method delivers outstanding performance on `Car' and `Cyclist', in which we gain a great improvement in 5 frame intervals (\textbf{9.4\%↑}/\textbf{10.3\%↑} for `Car' and \textbf{10.8\%↑}/\textbf{10.7\%↑} for `Cyclist') and 10 frame intervals (\textbf{16.4\%↑}/\textbf{19.6\%↑} for `Car' and \textbf{10.2\%↑}/\textbf{16.3\%↑} for `Cyclist'). However, the challenge of tracking large objects persists in high temporal variation cases for our method. Note that the performance of CXTrack drops dramatically after 3 frame intervals. In particular, in the medium variation case of 5 frame intervals, we achieve \textbf{11.3\%↑}/\textbf{15.7\%↑} improvement in overall success/precision compared to CXTrack, despite the fact that \emph{our HVTrack shares the same backbone and RPN with CXTrack}~\cite{xu2023cxtrack}. Furthermore, HVTrack surpasses CXTrack on `Car' and `Cyclist' by a very large margin (\textbf{21.7\%↑}/\textbf{26.7\%↑} for `Car' and \textbf{32.5\%↑}/\textbf{38.4\%↑} for `Cyclist'). The distinct performance gap between HVTrack and CXTrack in HV tracking showcases the effectiveness of our feature correlation module design.

\begin{table*}[ht]\footnotesize
  \centering
  \caption{Ablation analysis of HVTrack.}
  \resizebox{0.65\linewidth}{!}{
    \begin{tabular}{ccc|ccccc}
    \toprule[1.5pt]
    OM  & BEA & CPA & Car   & Pedestrian & Van   & Cyclist & Mean \\
    \midrule
          & \checkmark & \checkmark & 60.0/\underline{69.0} & 33.9/50.0 & \underline{28.4}/32.2 & \underline{54.2}/\underline{67.1} & 45.8/\underline{57.5} \\
    \checkmark &       & \checkmark & \textbf{60.3}/\textbf{69.4} & \underline{35.0}/\underline{50.2} & 26.7/30.7 & 43.9/61.5 & \underline{46.0}/\underline{57.5} \\
    \checkmark & \checkmark &       & 58.2/66.9 & 34.7/49.8 & 28.1/\textbf{33.5} & 47.7/63.9 & 45.1/56.5 \\
    \checkmark & \checkmark & \checkmark & \textbf{60.3}/68.9 & \textbf{35.1}/\textbf{52.1} & \textbf{28.7}/\underline{32.4} & \textbf{58.2}/\textbf{71.7} & \textbf{46.6}/\textbf{58.5} \\
    \bottomrule[1.5pt]
    \end{tabular}%
    }
  \label{tab:ablation}%
\end{table*}%

\begin{table}[ht]
  \centering
      \caption{Ablation experiment of BEA. `Base'/`Expansion' denotes only using the base/expansion branch in BEA.}
    \resizebox{0.6\linewidth}{!}{
    \begin{tabular}{c|ccccc}
    \toprule[1.5pt]
    Category & \multicolumn{1}{c}{Car} & \multicolumn{1}{c}{Pedestrian} & \multicolumn{1}{c}{Van} & \multicolumn{1}{c}{Cyclist} & \multicolumn{1}{c}{Mean} \\
    \midrule
    Base  & \multicolumn{1}{c}{60.3/69.4} & \multicolumn{1}{c}{35.0/50.2} & \multicolumn{1}{c}{26.7/30.7} & \multicolumn{1}{c}{43.9/61.5} & \multicolumn{1}{c}{46.0/57.5} \\
    Expansion & 60.0/68.6 & 34.7/50.5 & 31.4/36.8 & 54.5/67.5 & 46.4/57.9 \\
    \bottomrule[1.5pt]
    \end{tabular}%
    }
  \label{tab:BEA}%
\end{table}%

\noindent\textbf{Results on regular tracking.} For the KITTI dataset, we compare HVTrack with 12 top-performing trackers~\cite{giancola_leveraging_2019,qi_p2b_2020,fang_3d-siamrpn_2020,wang_mlvsnet_2021,hui_3d_2021,zhou2022pttr,zheng_box-aware_2021,shan_ptt_2021,hui_3d_2022,zheng_beyond_2022,xu2023cxtrack,lan2022temporal}. As shown in \cref{tab:KITTI}, our overall performance is close to the SOTA tracker CXTrack~\cite{xu2023cxtrack}, and achieves the second best result on the average in success  (2.0\%↓ w.r.t. CXTrack). Note that HVTrack outperforms TAT~\cite{lan2022temporal} on average (0.8\%↑/0.3\%↑), which utilizes temporal information by concatenating historical template features. This demonstrates our better design for leveraging the spatio-temporal context information. However, the performance of HVTrack drops when dealing with large objects (`Van'). We conjecture this performance drop to be caused by CPA, which will be further explored in \cref{sec:analysis}.
For the Waymo dataset, following the benchmark setting in LiDAR-SOT~\cite{pang2021model} and STNet~\cite{hui_3d_2022}, we test our HVTrack in 2 categories (`Vehicle', `Pedestrian') with 3 difficulty levels. All the methods are pre-trained on KITTI. The results of P2B~\cite{qi_p2b_2020}, BAT~\cite{zheng_box-aware_2021}, and V2B~\cite{hui_3d_2021} on Waymo are provided by STNet~\cite{hui_3d_2022}. As shown in \cref{tab:Waymo}, our method achieves the best performance in success (0.8\%↑) and precision (1.4\%↑). Notably, HVTrack does not surpass CXTrack and reach SOTA on the KTTTI benchmark, while the opposite situation occurs in the larger dataset of Waymo. The improvement on Waymo clearly demonstrates the robustness of our method in the large-scale dataset. Also, HVTrack surpasses other SOTA methods on all categories of `Hard' difficulty, revealing our excellent ability to handle sparse cases. \emph{The experimental results show that our method can generally solve the problem of 3D SOT under various levels of point cloud variations, and achieve outstanding performance.}

\subsection{Analysis Experiments}
\label{sec:analysis}

In this section, we extensively analyze HVTrack via a series of experiments. All the experiments are conducted on KITTI-HV with 5 frame intervals unless otherwise stated.

\noindent\textbf{Ablation Study.} We conduct experiments to analyze the effectiveness of different modules in HVTrack. As shown in \cref{tab:ablation}, we respectively ablate OM, BEA, and CPA from HVTrack. We only ablate OM in RPM because LM and MM serve as the template and are the indivisible parts of HVTrack. BEA and CPA are replaced by vanilla cross-attention and self-attention. In general, all components have been proven to be effective; removing an arbitrary module degrades the `mean’ performance. 

\noindent\textbf{Analysis Experiment of BEA.} The performance slightly drops on the `Car' when we apply BEA on HVTrack as shown in \cref{tab:ablation}. We conjecture this to be caused by the side effect of aggregating larger scale features in BEA, which will involve more background noise at each point. Further, `Car' has a medium size and does not have the distraction of crowded similar objects like small objects (`Pedestrian' and `Cyclist'), nor does it require a larger receptive field like large objects (`Van'). To verify this issue, we further analyze each branch of BEA as shown in \cref{tab:BEA}. `Pedestrian', `Van', and `Cyclist' benefit from the expansion branch and achieve a better performance compared to using only the base branch in BEA. On the other hand, the performance in the `Car' category has the opposite behavior to the other categories. The experimental results validate our hypothesis that BEA is beneficial to small and large objects, while negatively affecting medium-sized objects.

\begin{figure}[t]
  \centering
  \includegraphics[width=0.65\linewidth]{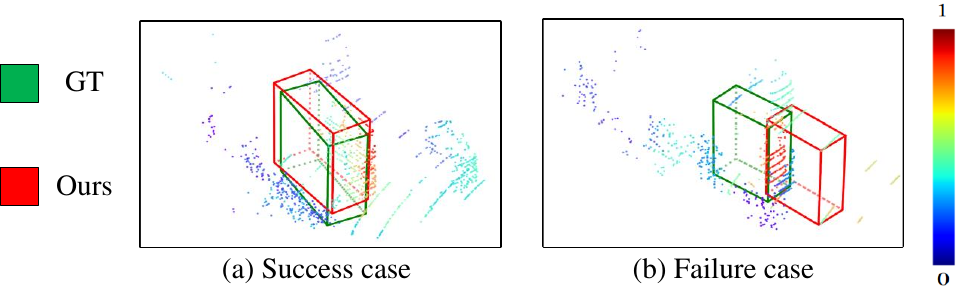}
    \caption{The attention maps of `Van' in CPA.}
  \label{fig:attn_map}
\end{figure}

\begin{table}[t]
  \centering
  \caption{\textbf{Results of HVTrack when using different memory sizes.} We train HVTrack with a memory size of $2$, and evaluate it with memory sizes ranging from $1$ to $8$ .}
  \resizebox{0.65\linewidth}{!}{
    \begin{tabular}{c|ccccc}
    \toprule[1.5pt]
    Memory Size & Car   & Pedestrian & Van   & Cyclist & Mean \\
    \midrule
    1     & 58.3/66.5 & 30.9/46.2 & 26.8/29.8 & 57.1/70.5 & 43.6/54.6 \\
    2     & 58.6/67.0 & 31.7/47.9 & 27.1/30.6 & 57.6/70.9 & 44.1/55.6 \\
    3     & 59.2/67.6 & 33.8/49.9 & 27.7/31 & 55.8/67.7 & 45.3/56.7 \\
    4     & \underline{60.0}/\underline{68.5} & 33.7/50.6 & \textbf{29.5}/\textbf{33.6} & 57.9/71.3 & 45.9/57.7 \\
    5     & \underline{60.0}/\underline{68.5} & 33.8/51.2 & \underline{28.7}/\underline{32.6} & 57.8/70.8 & 45.8/57.9 \\
    6     & \textbf{60.3}/\textbf{68.9} & \underline{35.1}/52.1 & \underline{28.7}/32.4 & \textbf{58.2}/\textbf{71.7} & \textbf{46.6}/\textbf{58.5} \\
    7     & 59.7/68.2 & \textbf{35.6}/\textbf{52.9} & 28.0/31.5 & \underline{58.1}/\underline{71.4} & \underline{46.4}/\underline{58.4} \\
    8     & 59.8/68.3 & \underline{35.1}/\underline{52.4} & 28.2/32.0 & \underline{58.1}/\underline{71.4} & 46.3/58.3 \\
    \bottomrule[1.5pt]
    \end{tabular}%
    }
  \label{tab:memory_size}%
\end{table}%

\noindent\textbf{Analysis Experiment of CPA.} Our method yields better results on `Van' after we remove CPA as shown in \cref{tab:ablation}, which reveals the relation between CPA and the large object tracking challenge.  We believe that this is caused by the suppressing strategy in CPA. Large objects usually have more points, and under the same probability of misclassification of importance, they will have more foreground points assigned as low importance in the attention map, resulting in a part of useful information being suppressed in CPA. As shown in \cref{fig:attn_map}b, the importance conflict in the object leads to tracking failure. That part of the information will be further suppressed when stacking multiple transformer layers. However, the performance drops in other categories, without CPA to suppress the background noise for medium and small objects. As shown in \cref{fig:attn_map}a, most of the background points are assigned with low importance and suppressed in the success case, which proves our idea of CPA.

\noindent\textbf{Memory Size.} Intuitively, trackers will achieve better performance when leveraging more temporal information. However, the performance of the trackers cannot continuously improve with the accumulation of historical information, due to inaccuracies in the historical tracklets. As shown in \cref{tab:memory_size}, we train HVTrack with a memory size of 2 due to the GPU memory limitation, and evaluate it with memory sizes from 1 to 8. The performance peaks for a memory size of 6, which is consistent with our assumption. Thus, we set 6 as our memory size and achieve a tracking speed of 31 FPS.

\section{Conclusion}
In this paper, we have explored a new task in 3D SOT, and presented the first 3D SOT framework for high temporal variation scenarios, HVTrack. Its three main components, RPM, BEA, and CPA, allow HVTrack to achieve robustness to point cloud variations, similar object distractions, and background noise. Our experiments have demonstrated that HVTrack significantly outperforms the state of the art in high temporal variation scenarios, and achieves remarkable performance in regular tracking. 

\noindent\textbf{Limitation.} Our CPA relies on fixed manual hyperparameters to suppress noise. This makes it difficult to balance the performance in different object and search area sizes, leading to a performance drop in tracking large objects. In the future, we will therefore explore the use of a learnable function to replace the manual hyperparameters in CPA and overcome the large object tracking challenge.

\section*{Acknowledgements}
This work is supported in part by the National Natural Science Foundation of China (NFSC) under Grants 62372377 and 62176242.

%
%
\bibliographystyle{splncs04}
\bibliography{main}

\clearpage
\input{supplementary}

\clearpage
\end{document}

%% file: supplementary.tex
\clearpage
\setcounter{page}{1}
\appendix
\label{sec:appendix}








\section{Implementation Details}
\label{sec:ImmpleDetail}

\noindent\textbf{KITTI-HV.} KITTI-HV has the same size as the original KITTI. We can simply construct KITTI-HV with a few lines of code as in \cref{alg:code}. We set the intervals non-linearly ([2,3,5,10]) instead of the traditional linear setting ([2,4,6,8]). Thus, we have denser tests in point cloud variations close to smooth scenarios (comparing [2,3,5] to [2,4,6]) for a fairer comparison with the existing methods.

\begin{algorithm}[htbp]
\caption{KITTI-HV Pseudocode, Python-like}
\label{alg:code}
\definecolor{codeblue}{rgb}{0.25,0.5,0.5}
\definecolor{codekw}{rgb}{0.85, 0.18, 0.50}
\lstset{
  backgroundcolor=\color{white},
  basicstyle=\fontsize{8.5pt}{8.5pt}\ttfamily\selectfont,
  columns=fullflexible,
  breaklines=true,
  captionpos=b,
  commentstyle=\fontsize{8.5pt}{8.5pt}\color{codeblue},
  keywordstyle=\fontsize{8.5pt}{8.5pt}\color{codekw},
}
\begin{lstlisting}[language=python]
# HV-tracklets: tracklets in KITTI-HV
for tracklet in KITTI:  # read tracklets in KITTI
    for i in range(min(len(tracklet),interval)):
    # starting at different frame
        temp_tracklet = tracklet[i::interval] 
        # sampling at frame intervals
        HV-tracklets.append(temp_tracklet) 
return HV-tracklets
\end{lstlisting}
\end{algorithm}

\noindent\textbf{Search areas}. Former trackers~\cite{qi_p2b_2020,zheng_box-aware_2021,zheng_beyond_2022,xu2023cxtrack} determine the search area by enlarging the target bounding box in wide and length at the last frame by 2 meters offset. We follow their strategy to generate the search area with enlargement offsets on KITTI~\cite{geiger_are_2012} as shown in \cref{tab:search_area}. We first statistically analyze the moving distance in the xy-plane of `Car' on KITTI with different frame intervals as shown in \cref{tab:static_analy}. We evaluate the performance of BAT~\cite{zheng_box-aware_2021} and M2-Track~\cite{zheng_beyond_2022} with different bounding box enlargement offsets in 5 frame intervals on KITTI-HV. The enlargement offsets are generated by slightly increasing the moving distances under different quantiles in \cref{tab:static_analy}. As illustrated in \cref{tab:trending}, BAT and M2-Track reach the peak at the enlargement offset of 4 meters and 6 meters, respectively. Thus, we choose the moving distances between quantiles of $50\%$ and $75\%$ as the enlargement offset for all the frame intervals and categories. Following~\cite{qi_p2b_2020,zheng_box-aware_2021,zheng_beyond_2022,xu2023cxtrack}, we randomly sample 1024 points in the search area as the input of the backbone.

\noindent\textbf{Observation angle}. Instead of the original radian $\in \mathbb{R}^{1}$, we use the sine and cosine values $\in \mathbb{R}^{2}$ to represent the observation angle.

\begin{figure*}[htbp]
  \centering
  \includegraphics[width=0.9\linewidth]{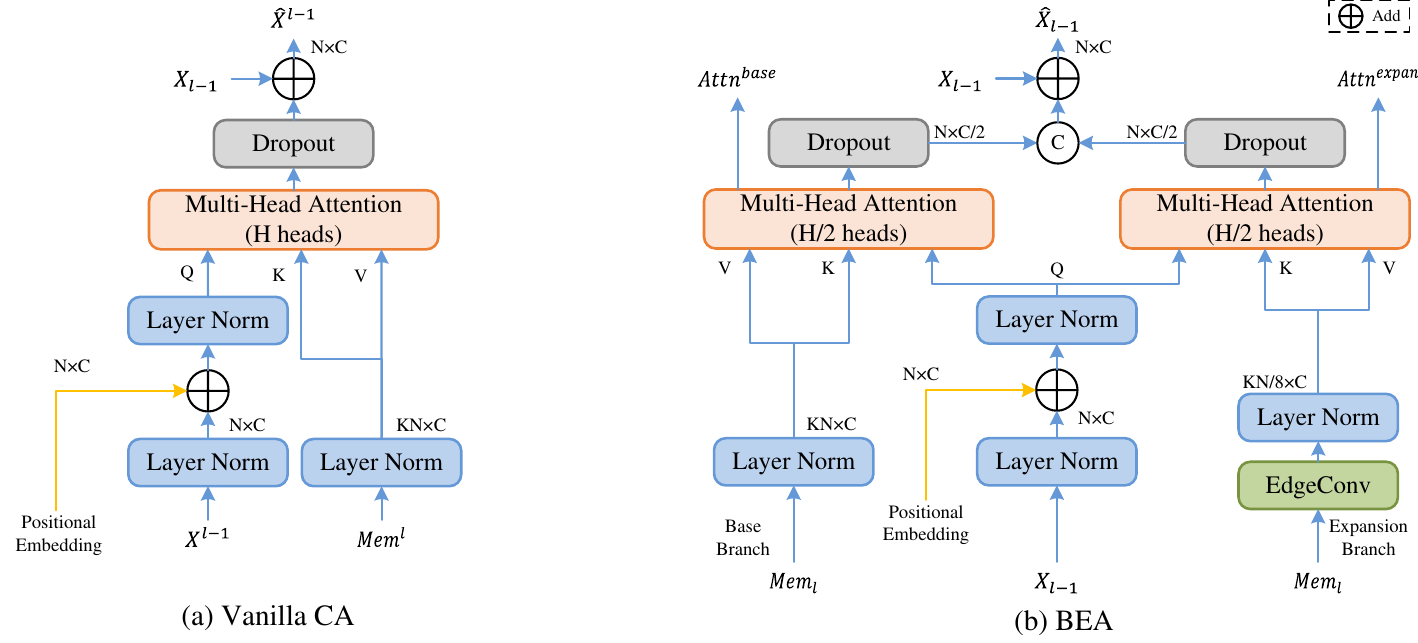}
    \caption{(a) Vanilla Cross-Attention (CA) and (b) Base-Expansion Feature Cross-Attention (BEA).}
  \label{fig:BEA_wide}
\end{figure*}

\begin{figure*}[htbp]
  \centering
  \includegraphics[width=0.9\linewidth]{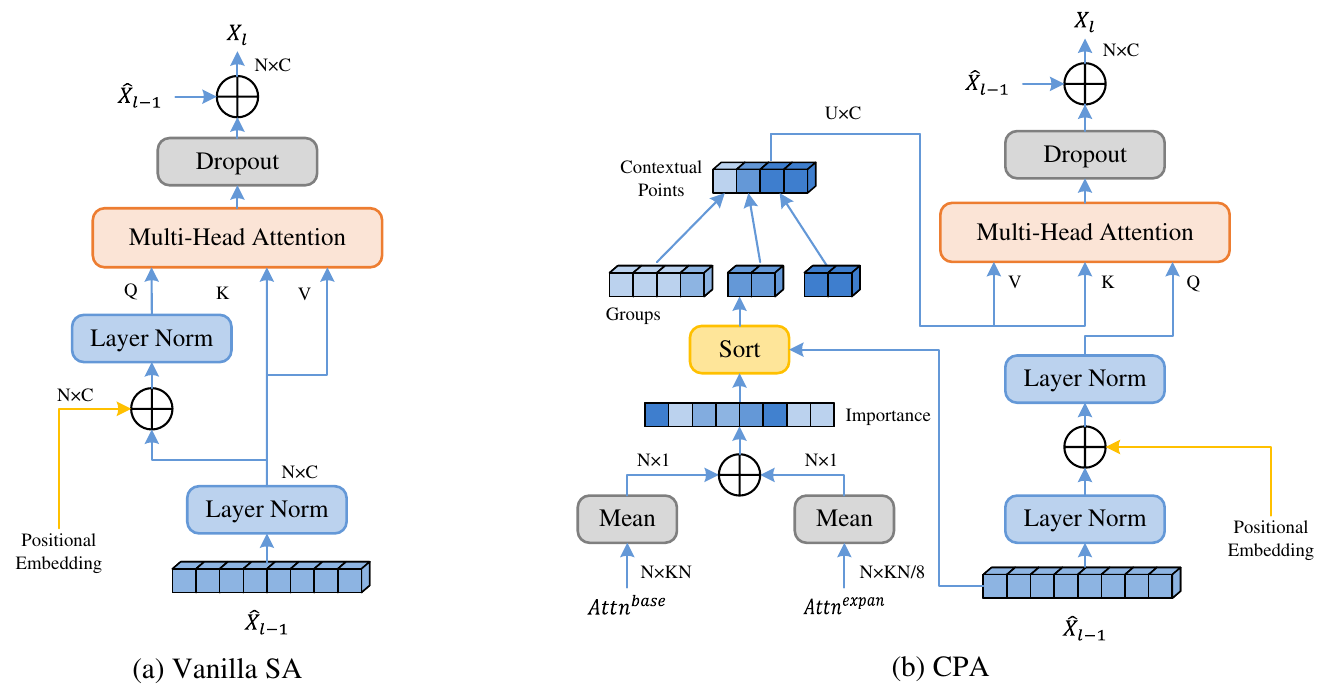}
    \caption{(a) Vanilla Self-Attention (CA) and (b) Contextual Point Guided Self-Attention (CPA).}
  \label{fig:CPA_wide}
\end{figure*}

\noindent\textbf{Ablation details}. We construct the vanilla cross-attention and self-attention in the ablation experiment as shown in \cref{fig:BEA_wide} (a) and \cref{fig:CPA_wide} (a). Compared to the BEA, vanilla cross-attention removes the expansion branch and assigns $H$ heads for the base branch. For the vanilla self-attention, we directly project $\hat{X}_{l-1}$ to K and V.
\begin{table}[htbp]\small
  \centering
  \caption{Bounding box enlargement offsets (meter) in different frame intervals and categories on KITTI for generating search areas.}
  \resizebox{0.42\linewidth}{!}{
    \begin{tabular}{c|cccc}
    \toprule[1.5pt]
    Frame Intervals & Car  & Pedestrian & Van  & Cyclist \\
    \midrule
    1     & 2     & 2     & 2     & 2 \\
    2     & 2     & 2     & 3     & 2 \\
    3     & 3     & 2     & 3     & 2 \\
    5     & 4     & 2     & 5     & 3 \\
    10    & 7     & 3     & 8     & 4 \\
    \bottomrule[1.5pt]
    \end{tabular}%
    }
  
  \label{tab:search_area}%
\end{table}%

\begin{table}[htbp]\small
  \centering
  \caption{Quantiles of Car's moving distance in the xy-plane with different frame intervals on the training set of KITTI.}
  \resizebox{0.4\linewidth}{!}{
    \begin{tabular}{c|ccccc}
    \toprule[1.5pt]
    Quantile & 1     & 2     & 3     & 5     & 10 \\
    \midrule
    25\% & 0.32  & 0.52  & 0.57  & 0.44  & 0.00 \\
    50\% & 0.79  & 1.55  & 2.28  & 3.61  & 5.51 \\
    75\% & 1.07  & 2.11  & 3.12  & 5.07  & 9.28 \\
    95\% & 2.26  & 4.38  & 6.06  & 9.17  & 15.38 \\
    99.73\% & 3.46  & 6.90   & 10.30  & 17.07 & 32.88 \\
    100\% & 14.56 & 15.48 & 16.49 & 19.21 & 36.56 \\
    \midrule
    Average & 0.81  & 1.57  & 2.28  & 3.53  & 5.78 \\
    \bottomrule[1.5pt]
    \end{tabular}%
    }
  
  \label{tab:static_analy}%
\end{table}%

\begin{table*}[htbp]
  \centering
  \caption{Performance of BAT and M2-Track in different search area sizes on `Car' of KITTI-HV with 5 frame intervals. We determine the search area size by enlarging the object bounding box in width and length with an offset.}
  \resizebox{0.75\linewidth}{!}{
    \begin{tabular}{c|cc|cc|cc|cc|cc}
    \toprule[1.5pt]
    Offset (m) & \multicolumn{2}{c|}{20} & \multicolumn{2}{c|}{18} & \multicolumn{2}{c|}{10} & \multicolumn{2}{c|}{6} & \multicolumn{2}{c}{4} \\
    \midrule
    Method & Succ.  & Prec.  & Succ.  & Prec.  & Succ.  & Prec.  & Succ.  & Prec.  & Succ. & Prec. \\
    \midrule
    BAT~\cite{zheng_box-aware_2021}   & 16.62 & 16.88 & 17.02 & 17.18 & 25.27 & 27.70  & 35.05 & 40.25 & {\textbf{44.13}} & {\textbf{51.11}} \\
    M2-Track~\cite{zheng_beyond_2022} & 16.53 & 14.59 & 21.69 & 22.51 & 43.12 & 50.80  & {\textbf{52.64}} & {\textbf{61.58}} & 50.87 & 58.56 \\
    \bottomrule[1.5pt]
    \end{tabular}%
    }
  \label{tab:trending}%
\end{table*}%

\section{More Comparisons}

\noindent\textbf{Comparison with latest SOTAs.} In \cref{tab:KITTI_recent_sota}, we compare HVTrack with the latest SOTAs on KITTI. There still exists a performance gap compared to them. M3SOT~\cite{liu2024m3sot} extends MBPTrack~\cite{xu2023mbptrack} via the SpaceFormer and achieves better performance. Thus, we report the stronger tracker M3SOT in high temporal variation scenarios in \cref{tab:KITTIHV_recent_sota} to validate the effectiveness of HVTrack. HVTrack still yields the best results at various intervals, with a notable improvement of 17.2\%/21.3\% at 5 intervals.

\noindent\textbf{Efficiency.} We compare HVTrack with SOTA methods in efficiency on KITTI-HV with 5 frame intervals in \cref{tab:running_time}. Due to the increased search area, CXTrack shows a 26.5\% speed decline compared to the 34 FPS reported in its paper. 

\noindent\textbf{Backbone flexibility}. As illustrated in \cref{tab:abla_backbone}, we conduct analysis experiments using different backbones in HVTrack on KITTI-HV with 5 frame intervals. PointNet++~\cite{qi2017pointnet++} is widely used in former trackers~\cite{giancola_leveraging_2019,qi_p2b_2020,hui_3d_2021,hui_3d_2022,wang_mlvsnet_2021,zheng_box-aware_2021,shan_ptt_2021,zheng_beyond_2022,zhou2022pttr,guo2022cmt,cui20213d}, and GCDNN~\cite{wang2019dynamic} is employed in~\cite{xu2023cxtrack}. Our HVTrack shows robust performance with different backbones, demonstrating the strong flexibility of our approach. In particular, HVTrack achieves an improvement with 0.7\%↑/1.5\%↑ on the average in success/precision, confirming the great potential for further improvement.

\noindent\textbf{One pre-trained model.} We report the results of KITTI pre-trained models on KITTI-HV in \cref{tab:one_pretrained} (top). Our memory module requires rich object pose samples to fit object motion. Thus HVTrack suffers a performance degradation on `Car'.
However, the performance improvement on `Pedestrian' proves the effectiveness of HVTrack when the object pose distribution changes only slightly. To fully demonstrate the generalizability of HVTrack, we train models in [1,2,3,5,10] intervals together, and test them under different intervals in \cref{tab:one_pretrained} (bottom). In contrast to other methods whose performance decreases as the interval grows, HVTrack maintains consistent performance across [1,2,3,5] intervals. This demonstrates the robustness of our method in different temporal variation scenarios.

\noindent\textbf{Waymo-HV}. Following the construction of KITTI-HV, we build Waymo-HV for a more comprehensive comparison as illustrated in \cref{tab:waymo_hv}. Our HVTrack consistently outperforms the state-of-the-art methods~\cite{zheng_box-aware_2021,xu2023cxtrack} across all frame intervals.

\begin{table}[t]

  \centering
  \caption{Comparison with the most recent SOTAs on KITTI.}
  
  \resizebox{0.8\linewidth}{!}{
    \begin{tabular}{c|c|c|c|c|c|c}
    \toprule[1.5pt]
    Category & Car   & Pedestrian & Van   & Cyclist & Mean & Params (MB)\\
    \midrule
    MBPTrack~\cite{xu2023mbptrack} & 73.4/84.8 & \textbf{68.6}/\textbf{93.9} & \textbf{61.3}/72.7 & \textbf{76.7}/\textbf{94.3} & 70.3/87.9 & 7.39\\
    M3SOT~\cite{liu2024m3sot} & \textbf{75.9}/\textbf{87.4} & 66.6/92.5 & 59.4/\textbf{74.7} & 70.3/93.4 & \textbf{70.3}/\textbf{88.6} & 16.43 \\
    \midrule
    HVTrack & 68.2/79.2 & 64.6/90.6 & 54.8/63.8 & 72.4/93.7 & 65.5/83.1 & \textbf{5.60} \\
    \bottomrule[1.5pt]
    \end{tabular}%
    }
  \label{tab:KITTI_recent_sota}%
  
\end{table}%

\begin{table}[t]

  \centering
  \caption{Comparison with the most recent SOTA on KITTI-HV.}
  
  \resizebox{0.8\linewidth}{!}{
    \begin{tabular}{c|c|c|c|c|c|c}
    \toprule[1.5pt]
    \multicolumn{1}{c|}{Interval} & Method & Car   & Pedestrian & Van   & Cyclist & Mean \\
    \midrule
    \multirow{2}[6]{*}{2} & M3SOT~\cite{liu2024m3sot} & 59.0/67.9 & \textbf{61.7}/\textbf{86.3}  & \textbf{55.2}/\textbf{68.7} & 55.1/86.3 & 59.8/76.3 \\
\cmidrule{2-7}          & HVTrack & \textbf{67.1}/\textbf{77.5} & 60.0/84.0 & 50.6/61.7 & \textbf{73.9}/\textbf{93.6} & \textbf{62.7}/\textbf{79.3} \\
    \midrule
    \midrule
    \multirow{2}[6]{*}{3} & M3SOT~\cite{liu2024m3sot} & 46.9/52.6 & 50.1/\textbf{74.0} & \textbf{43.3}/\textbf{53.7} & 32.4/48.1 & 47.7/61.9 \\
\cmidrule{2-7}          & HVTrack & \textbf{66.8}/\textbf{76.5} & \textbf{51.1}/71.9 & 38.7/46.9 & \textbf{66.5}/\textbf{89.7} & \textbf{57.5}/\textbf{72.2} \\
    \midrule
    \midrule
    \multirow{2}[6]{*}{5} & M3SOT~\cite{liu2024m3sot} & 30.5/34.5 & 31.0/44.0 & 18.3/21.0 & 21.6/25.9 & 29.4/37.2 \\
\cmidrule{2-7}          & HVTrack & \textbf{60.3}/\textbf{68.9} & \textbf{35.1}/\textbf{52.1} & \textbf{28.7}/\textbf{32.4} & \textbf{58.2}/\textbf{71.7} & \textbf{46.6}/\textbf{58.5} \\
    \midrule
    \midrule
    \multirow{2}[6]{*}{10} & M3SOT~\cite{liu2024m3sot} & 26.1/26.6 & 16.2/18.8 & 17.6/17.1 & 27.5/26.2 & 21.1/22.4 \\
\cmidrule{2-7}          & HVTrack & \textbf{49.4}/\textbf{54.7} & \textbf{22.5}/\textbf{29.1} & \textbf{22.2}/\textbf{23.4} & \textbf{39.5}/\textbf{45.4} & \textbf{35.1}/\textbf{40.6} \\
    \bottomrule[1.5pt]
    \end{tabular}%
    }
  \label{tab:KITTIHV_recent_sota}%
  
\end{table}%

\begin{table}[t]

  \centering
  \caption{Comparison in efficiency with SOTA.}
  \resizebox{0.7\linewidth}{!}{
    \begin{tabular}{c|cccc}
    \toprule[1.5pt]
    Method & M2-Track~\cite{zheng_beyond_2022} & CXTrack~\cite{xu2023cxtrack} & M3SOT~\cite{liu2024m3sot} & HVTrack \\
    \midrule
    FPS   & \textbf{42}    & 25    & 14    & \underline{31} \\
    Params (MB) & \underline{8.54} & 18.27 & 16.43 & \textbf{5.60} \\
    \bottomrule[1.5pt]
    \end{tabular}%
    }
  \label{tab:running_time}%
  
\end{table}%

\begin{table*}[t]
  \centering
  \caption{Analysis experiments of using different backbones in HVTrack on KITTI-HV with 5 frame intervals.}
    \begin{tabular}{c|ccccc}
    \toprule[1.5pt]
    Category & Car   & Pestrian & Van   & Cyclist & Mean \\
    Frame Number & 6424  & 6088  & 1248  & 308   & 14068 \\
    \midrule
    DGCNN~\cite{wang2019dynamic}& \textbf{60.3}/\textbf{68.9} & 35.1/52.1 & \textbf{28.7}/\textbf{32.4} & \textbf{58.2}/\textbf{71.7} & 46.6/58.5 \\
    PointNet++~\cite{qi2017pointnet++} & 58.6/66.7 & \textbf{39.0}/\textbf{58.3} & 27.5/30.7 & 57.4/70.9 & \textbf{47.3}/\textbf{60.0} \\
    \bottomrule[1.5pt]
    \end{tabular}%
  \label{tab:abla_backbone}%
\end{table*}%

\begin{table}[htbp]

  \centering
  \caption{Comparison of different training settings on KITTI-HV.}
  
  \resizebox{0.95\linewidth}{!}{
    \begin{tabular}{c|c|c|ccccc}
    \toprule[1.5pt]
    \multicolumn{1}{c|}{Training} & \multirow{2}[2]{*}{Category} & \multirow{2}[2]{*}{Method} & \multicolumn{5}{c}{Testing interval} \\
    interval(s) &       &       & 1     & 2     & 3     & 5     & 10 \\
    \midrule
    \multirow{6}[4]{*}{1} & \multirow{3}[2]{*}{Car} & M2-Track & 65.5/80.8 & \textbf{62.8}/\textbf{74.4} & \textbf{52.5}/\textbf{61.0} & \textbf{36.1}/\textbf{39.8} & \textbf{23.5}/\textbf{24.5} \\
          &       & CXTrack & \textbf{69.1}/\textbf{81.6} & 59.4/69.4 & 51.5/58.4 & 33.6/36.0 & 22.5/21.3 \\
          &       & HVTrack & 68.2/79.2 & 59.8/68.2 & 45.8/51.1 & 21.2/20.8 & 18.3/20.2 \\
\cmidrule{2-8}          & \multirow{3}[2]{*}{Pedestrian} & M2-Track & 61.5/88.2 & 58.7/86.5 & 50.8/74.4 & 30.7/42.3 & 16.3/19.5 \\
          &       & CXTrack & \textbf{67.0}/\textbf{91.5} & \textbf{64.9}/\textbf{88.0} & 56.4/78.7 & 36.2/48.0 & \textbf{18.3}/\textbf{21.2} \\
          &       & HVTrack & 64.6/90.6 & 63.6/87.8 & \textbf{60.5}/\textbf{82.6} & \textbf{42.7}/\textbf{57.6} & 16.9/19.6 \\
    \midrule
    \midrule
    \multicolumn{1}{c|}{\multirow{6}[4]{*}{1,2,3,5,10}} & \multirow{3}[2]{*}{Car} & M2-Track & 57.8/74.2 & 60.3/\textbf{73.7} & 57.1/66.7 & 59.9/68.8 & 37.5/40.0 \\
          &       & CXTrack & 57.8/70.2 & 51.5/60.3 & 52.2/58.3 & 34.9/38.3 & 25.1/24.6 \\
          &       & HVTrack & \textbf{65.6}/\textbf{76.5} & \textbf{60.3}/69.8 & \textbf{64.6}/\textbf{73.4} & \textbf{63.9}/\textbf{71.8} & \textbf{40.9}/\textbf{44.3} \\
\cmidrule{2-8}          & \multirow{3}[2]{*}{Pedestrian} & M2-Track & 53.0/79.2 & 49.3/70.6 & 41.9/60.9 & 37.0/54.7 & 24.0/30.9 \\
          &       & CXTrack & \textbf{60.3}/\textbf{84.4} & \textbf{60.1}/\textbf{84.5} & 52.8/73.7 & 33.2/44.1 & 17.2/19.6 \\
          &       & HVTrack & 56.4/78.9 & 58.5/81.2 & \textbf{58.2}/\textbf{79.7} & \textbf{56.3}/\textbf{77.2} & \textbf{30.6}/\textbf{39.1} \\
    \bottomrule[1.5pt]
    \end{tabular}%
    }
  \label{tab:one_pretrained}%
    
\end{table}%

\begin{table}[htbp]
  \centering
  \caption{Comparison of HVTrack with the state-of-the-art methods on each category of the Waymo-HV dataset.}
  \resizebox{1.0\linewidth}{!}{
    \begin{tabular}{c|c|c|c|c|c|c|c|c|c|c}
    \toprule[1.5pt]
    \multicolumn{1}{c|}{\multirow{2}[2]{*}{\makecell[c]{Frame\\Interval}}} & \multirow{2}[2]{*}{Method} & \multicolumn{4}{c|}{Vehicle (185632)} & \multicolumn{4}{c|}{Pedestrian (241168)} & \multirow{2}[2]{*}{Mean (426800)} \\
          &       & \multicolumn{1}{c}{Easy} & \multicolumn{1}{c}{Medium} & \multicolumn{1}{c}{Hard} & Mean  & \multicolumn{1}{c}{Easy} & \multicolumn{1}{c}{Medium} & \multicolumn{1}{c}{Hard} & Mean  &  \\
    \midrule
    \multirow{3}[4]{*}{2} & BAT~\cite{zheng_box-aware_2021}   & 61.0/68.3 & 53.3/60.9  & 48.9/57.8 & 54.7/62.7 & 19.3/32.6 & 17.8/29.8 & 17.2/28.3 & 18.2/30.3 & 34.1/44.4 \\
          & CXTrack~\cite{xu2023cxtrack} & 63.9/71.1 & 54.2/62.7 & 52.1/63.7 & 57.1/66.1 & \textbf{35.4}/\textbf{55.3} & \textbf{29.7}/\textbf{47.9} & 26.3/44.4 & \textbf{30.7}/\textbf{49.4} & 42.2/56.7 \\
\cmidrule{2-11}          & HVTrack(Ours) & \textbf{66.2}/\textbf{75.2} & \textbf{57.0}/\textbf{66.0} & \textbf{55.3}/\textbf{67.1} & \textbf{59.8}/\textbf{69.7} & 34.2/53.5 & 28.7/\textbf{47.9} & \textbf{26.7}/\textbf{45.2} & 30.0/49.1 & \textbf{43.0}/\textbf{58.1} \\
    \midrule
    \midrule
    \multirow{3}[4]{*}{3} & BAT~\cite{zheng_box-aware_2021}   & 47.1/52.3 & 39.8/45.2 & 35.1/40.6 & 41.0/46.4 & 18.2/27.4 & 15.4/22.8 & 13.7/19.8 & 15.9/23.5 & 26.8/33.5 \\
          & CXTrack~\cite{xu2023cxtrack} & 59.8/64.7 & 36.5/40.7 & 26.7/30.8 & 42.0/46.5 & \textbf{28.2}/\textbf{41.1} & \textbf{21.9}/\textbf{33.1} & \textbf{16.6}/\textbf{25.3} & \textbf{22.5}/\textbf{33.5} & 31.0/39.2 \\
\cmidrule{2-11}          & HVTrack(Ours) & \textbf{64.3}/\textbf{71.3} & \textbf{54.3}/\textbf{62.2} & \textbf{48.5}/\textbf{57.2} & \textbf{56.2}/\textbf{64.0} & 25.7/38.2 & 18.6/28.2 & 14.6/22.6 & 19.9/30.0 & \textbf{35.7}/\textbf{44.8} \\
    \midrule
    \midrule
    \multirow{3}[4]{*}{5} & BAT~\cite{zheng_box-aware_2021}   & 47.1/52.4 & 34.4/38.2 & 27.9/31.3 & 37.1/41.3 & 13.6/18.5 & 12.4/16.8 & 10.8/13.8 & 12.3/16.5 & 23.1/27.3 \\
          & CXTrack~\cite{xu2023cxtrack} & 45.9/50.5 & 27.1/29.2 & 19.5/21.1 & 31.7/34.6 & \textbf{23.0}/32.1 & \textbf{18.0}/\textbf{25.9} & \textbf{13.7}/\textbf{19.5} & \textbf{18.5}/\textbf{26.1} & 24.2/29.8 \\
\cmidrule{2-11}          & HVTrack(Ours) & \textbf{47.1}/\textbf{52.3} & \textbf{40.1}/\textbf{45.4} & \textbf{34.3}/\textbf{39.4} & \textbf{40.9}/\textbf{46.1} & 22.4/\textbf{32.2} & 17.5/25.5 & 13.5/19.3 & 18.0/26.0 & \textbf{28.0}/\textbf{34.7} \\
    \midrule
    \midrule
    \multirow{3}[4]{*}{10} & BAT   & 31.7/32.3 & 23.5/23.7 & 20.9/21.3 & 25.7/26.1 & 10.8/11.9 & 10.3/11.0 & 10.3/10.4 & 10.5/11.1 & 17.1/17.6 \\
          & CXTrack & 25.1/23.7 & 16.3/14.4 & 14.4/13.1 & 19.0/17.4 & 14.1/17.2 & 12.3/14.2 & 11.1/11.8 & 12.6/14.5 & 15.4/15.8 \\
\cmidrule{2-11}          & HVTrack(Ours) & \textbf{36.8}/\textbf{39.6} & \textbf{26.9}/\textbf{28.6} & \textbf{22.0}/\textbf{23.2} & \textbf{29.1}/\textbf{31.0} & \textbf{16.4}/\textbf{20.9} & \textbf{14.0}/\textbf{17.3} & \textbf{12.6}/\textbf{14.8} & \textbf{14.4}/\textbf{17.8} & \textbf{20.8}/\textbf{23.5} \\
    \bottomrule[1.5pt]
    \end{tabular}%
    }
  \label{tab:waymo_hv}%
\end{table}%

\begin{table*}[htbp]\footnotesize
  \centering
  \caption{\textbf{Comparison of HVTrack with the state-of-the-art methods on each category of the NuScenes dataset.}}
  {
    \begin{tabular}{c|c|c|c|c|c|c}
    \toprule
    Category & Car   & Pedestrian & Truck & Trailer & Bus   & Mean \\
    Frame Number & 64159 & 33227 & 13587 & 3352  & 2953  & 117278 \\
    \midrule
    SC3D~\cite{giancola_leveraging_2019}  & 22.3/21.9 & 11.3/12.7 & 30.7/27.7 & 35.3/28.1 & 29.4/24.1 & 20.7/20.2 \\
    P2B~\cite{qi_p2b_2020}   & 38.8/43.2 & 28.4/52.2 & 43.0/41.6 & 49.0/40.1 & 33.0/27.4 & 36.5/45.1 \\
    BAT~\cite{zheng_box-aware_2021}   & 40.7/44.3 & 28.8/53.3 & 45.3/42.6 & 52.6/44.9 & 35.4/28.0 & 38.1/45.7 \\
    M2-Track~\cite{zheng_beyond_2022} & \textbf{55.9}/\textbf{65.1} & \underline{32.1}/\underline{60.9} & \textbf{57.4}/\textbf{59.5} & \underline{57.6}/\textbf{58.3} & \textbf{51.4}/\textbf{51.4} & \underline{49.2}/\textbf{62.7} \\
    CXTrack~\cite{xu2023cxtrack} & 44.6/50.5 & 31.5/55.8 & 51.3/50.7 & \textbf{59.7}/\underline{53.6} & \underline{42.6}/37.3 & 42.0/51.8 \\
    \midrule
    HVTrack & \textbf{55.9}/\underline{62.9} & \textbf{41.3}/\textbf{67.6} & \underline{55.6}/55.2 & 52.0/40.2 & 36.3/41.6 & \textbf{51.1}/\underline{62.2} \\
    \bottomrule
    \end{tabular}%
    }
  
  \label{tab:nuscenes}%
  
\end{table*}%

\begin{table*}[htbp]
  \centering
  \caption{\textbf{Comparison of HVTrack with the state-of-the-art methods on each category of the nuScenes-HV dataset.} We construct the high-variation dataset nuScenes-HV for training and testing by setting 2 frame intervals for sampling in the NuScenes dataset.}
    \begin{tabular}{c|c|c|c|c|c|c}
    \toprule[1.5pt]
    Category & Car   & Pedestrian & Truck & Trailer & Bus   & Mean \\
    Frame Number & 64159 & 33227 & 13587 & 3352  & 2953  & 117278 \\
    \midrule
    P2B~\cite{qi_p2b_2020}   & 47.5/51.3 & 23.1/35.0 & 52.9/51.5 & 63.6/56.2 & 40.2/37.2 & 41.5/46.5 \\
    BAT~\cite{zheng_box-aware_2021}   & 44.7/48.0 & 23.1/33.2 & 52.3/50.9 & \underline{63.7}/\underline{57.7} & 41.6/38.2 & 39.9/44.2 \\
    M2-Track~\cite{zheng_beyond_2022} & \underline{51.7}/\underline{60.1} & \underline{37.8}/\underline{60.6} & \underline{55.4}/\textbf{57.8} & \textbf{65.8}/\textbf{64.8} & \textbf{51.5}/\textbf{49.2} & \underline{48.6}/\underline{59.8} \\
    CXTrack~\cite{xu2023cxtrack} & 50.7/57.6 & 27.0/43.8 & 54.3/55.0 & 62.2/56.5 & \underline{43.4}/\underline{40.5} & 44.5/52.9 \\
    \midrule
    HVTrack & \textbf{57.0}/\textbf{63.4} & \textbf{43.1}/\textbf{68.2} & \textbf{56.0}/\underline{56.1} & 51.7/43.1 & 31.2/35.2 & \textbf{52.4}/\textbf{62.6} \\
    \bottomrule[1.5pt]
    \end{tabular}%
  \label{tab:nuscenes_hard}%
\end{table*}%

\noindent\textbf{NuScenes}. Following the setting in M2-Track~\cite{zheng_beyond_2022}, we evaluate our HVTrack in 4 categories (`Car', `Truck', `Trailer' and `Bus') of the famous nuScenes~\cite{caesar_nuscenes_2019} dataset. The results of SC3D~\cite{giancola_leveraging_2019}, P2B~\cite{qi_p2b_2020}, and BAT~\cite{zheng_box-aware_2021} on NuScenes are provided by M2-Track. CXTrack~\cite{xu2023cxtrack} follows the dataset setting in STNet~\cite{hui_3d_2022}, which is quite different from M2-Track. We train CXTrack on NuScenes using its official code and report the results. As shown in \cref{tab:nuscenes}, our method achieves the best performance in success (1.9\%↑) and ranks second in precision (0.5\%↓). HVTrack surpasses M2-Track in `Pedestrian' with a great improvement in success (\textbf{9.2\%↑}) and precision (\textbf{6.6\%↑}), revealing our excellent ability to handle complex cases. `Pedestrian' is usually considered to have the largest point cloud variations and proportion of noise, due to the small object sizes and the diversity of body motion. Notably, we achieve \textbf{9.1\%↑/10.4\%↑} improvement in success/precision on average over CXTrack, which has the same backbone and RPN. This gap clearly demonstrates the robustness of our method in regular tracking. However, the performance of HVTrack still drops when dealing with large objects.

\noindent\textbf{NuScenes-HV}. As shown in \cref{tab:nuscenes_hard}, we compare HVTrack with the state-of-the-art methods on each category of the nuScenes-HV dataset. We construct the high-variation dataset nuScenes-HV for training and testing by setting 2 frame intervals for sampling in the NuScenes dataset. We achieve the best performance in both success (52.4\%, 3.8\%↑) and precision (62.6\%, 2.8\%↑) on average. We surpass SOTA trackers in the categories with a large number of samples (`Car', `Pedestrian', and `Truck'). However, our performance drops in `Trailer' and `Bus', which have a small number of samples. We believe the length of tracklets is another factor that affects the performance of HVTrack on `Trailer' and `Bus'. With 2 frame intervals, the average tracklet length of the `Trailer' is only 11.06 frames on nuScenes-HV, while it is 26.59 frames for the `Van' on KITTI-HV. With such a short average tracklet length, HVTrack is unable to obtain enough historical information for training and testing, leading to a performance drop. Further, a too short tracklet length is not in line with real-world scenarios. Therefore, we only construct nuScenes-HV with 2 frame intervals.

\section{Visualization Results}
\begin{figure*}[htbp]
  \centering
  \includegraphics[width=1.0\linewidth]{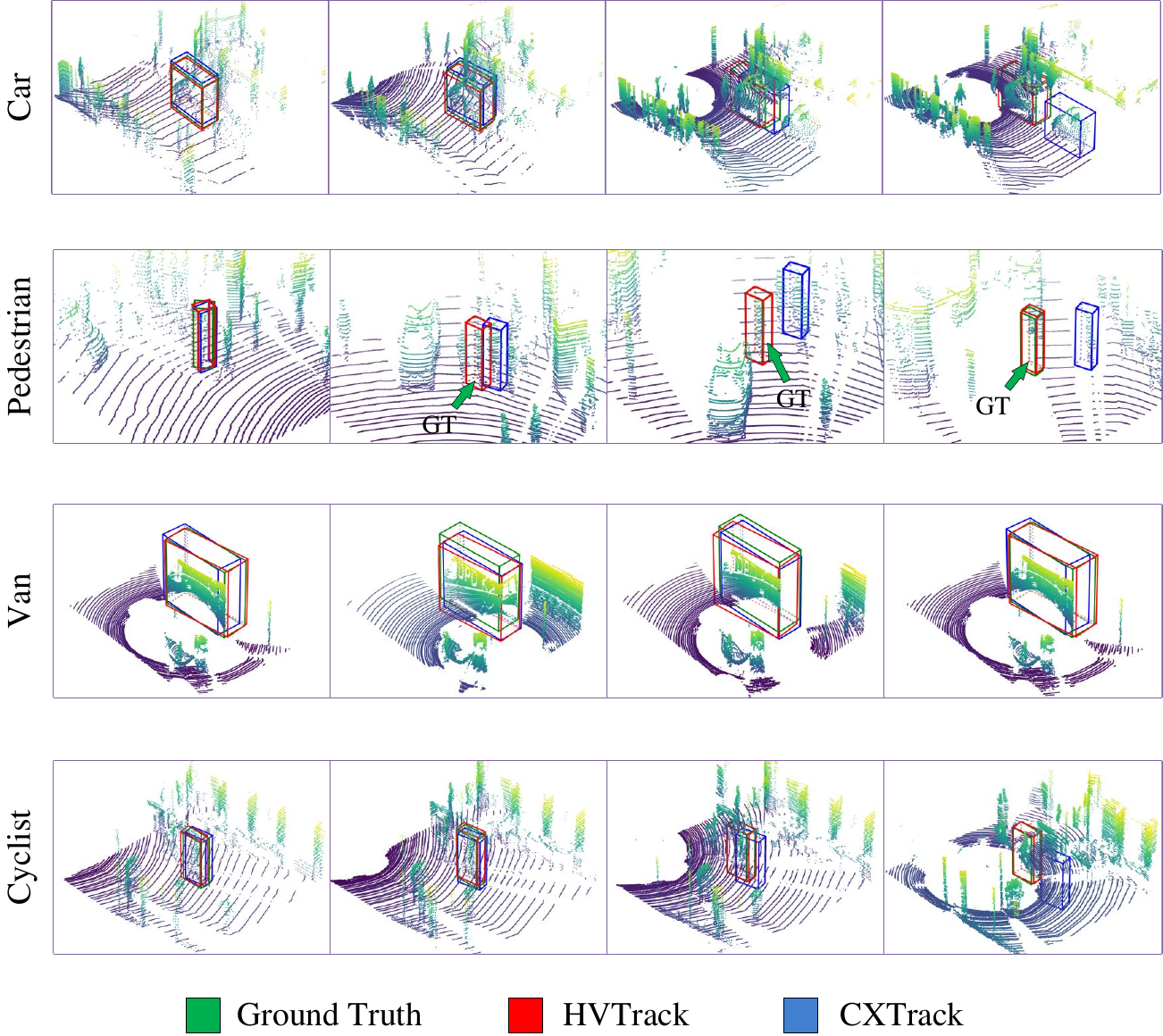}
    \caption{Visualization results in dense cases on KITTI-HV with 5 frame intervals.}
  \label{fig:vis_dense}
\end{figure*}

\begin{figure*}[htbp]
  \centering
  \includegraphics[width=1.0\linewidth]{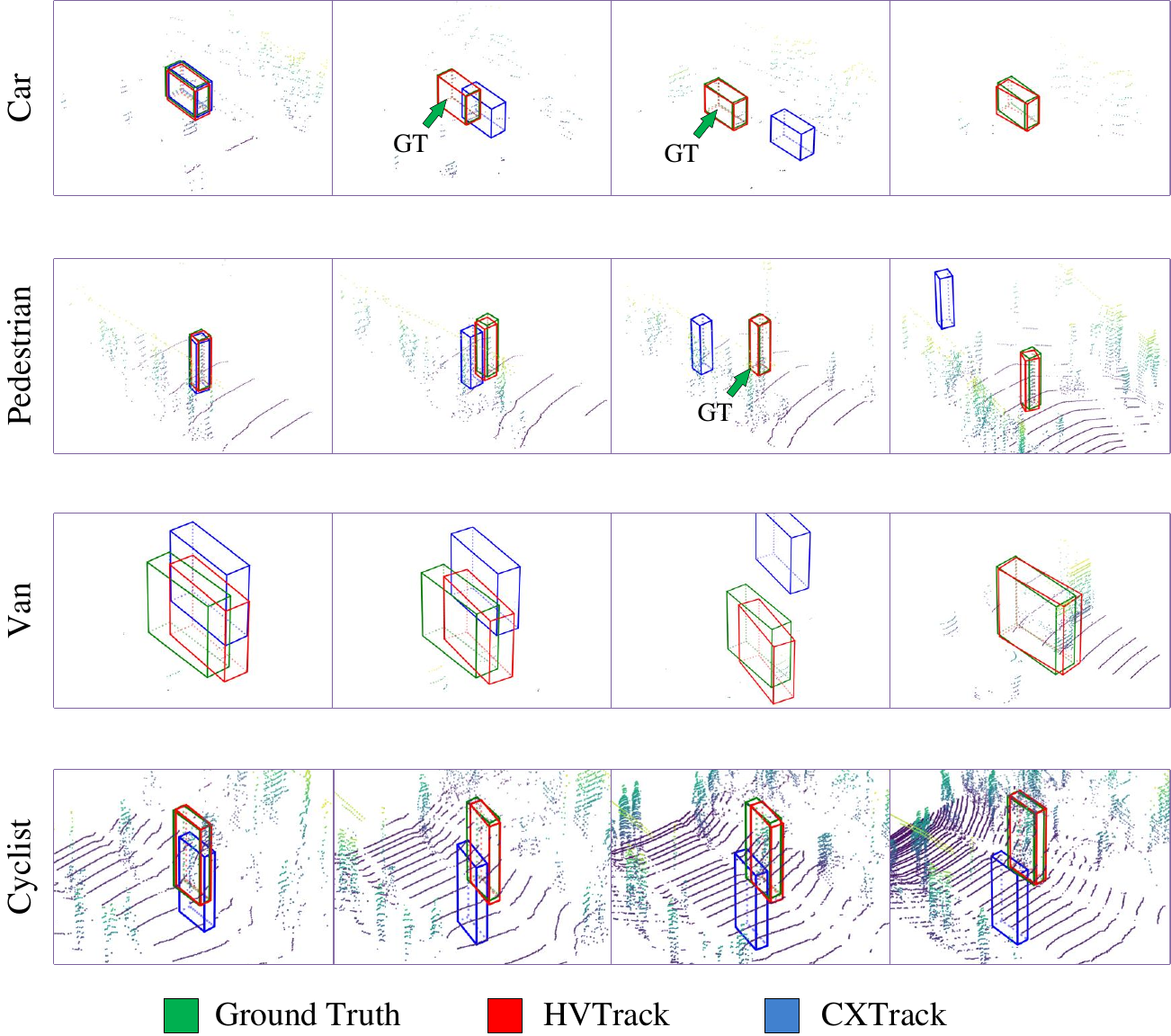}
    \caption{Visualization results in sparse cases on KITTI-HV with 5 frame intervals.}
  \label{fig:vis_sparse}
\end{figure*}

As illustrated in \cref{fig:vis_dense} and \cref{fig:vis_sparse}, we visualize our experiment results on KITTI-HV with 5 frame intervals in dense and sparse cases. The `Car', `Pedestrian', and `Cyclist' in \cref{fig:vis_dense} demonstrate the excellent performance of HVTrack in dealing with the distraction of similar objects and massive noise. Moreover, the success of the sparse cases in \cref{fig:vis_sparse} confirms the effective utilization of historical information in our method.

